\documentclass[journal, twocoloumn]{IEEEtran}
\usepackage{algorithm}
\usepackage{algpseudocode}
\usepackage{amsmath}
\DeclareMathOperator*{\argmax}{arg\,max}

\usepackage{graphicx}
\usepackage{multicol}
\usepackage{epstopdf}
\usepackage{cite}
\usepackage{float}

\usepackage{enumitem}
\usepackage{array}
\usepackage{lipsum}
\usepackage{fixltx2e}
\usepackage{amsfonts}
\usepackage{algpseudocode}%
\usepackage[numbers,square, comma, sort & compress]{natbib}
\ifCLASSOPTIONcompsoc
\usepackage[caption=true,font=normalsize,labelfont=sf,textfont=sf]{subfig}
\else
\usepackage[caption=true,font=footnotesize]{subfig}
\fi

\ifCLASSINFOpdf
\else
\fi
\usepackage{amsmath}
\usepackage[cmintegrals]{newtxmath}
\usepackage[T1]{fontenc}
\usepackage{lipsum}
\usepackage{bbm}
\usepackage{caption}
\usepackage{algorithm}
\usepackage{algpseudocode}
\usepackage{mathtools}

\newtheorem{thm}{Theorem}
\newtheorem{lemma}{Lemma}

\newtheorem{defn}{Definition}

\newtheorem{remark}{Remark}
\usepackage{dblfloatfix}
\usepackage{array}
\usepackage{color}

\newcommand{\de}{$\rm{dE}^3$}
\newcommand{\dets}{$\rm{dE}^3$-$\rm{TS}$}

\begin{document}

\title{Multi-player Bandits for Optimal Assignment with Heterogeneous Rewards}

\author{$\text{Harshvardhan Tibrewal}^{\dagger}$, $\text{Sravan Patchala}^{\dagger}$, Manjesh K. Hanawal and Sumit J. Darak
\thanks{Harshvardhan Tibrewal and Sravan Patchala are with the Electrical Engineering Department, IIT Bombay, India. Manjesh K. Hanawal is with Industrial Engineering and Operations Research Department, IIT Bombay, India. Sumit J. Darak is with the Electronics and Communications Department, IIIT Delhi, India.}}

\maketitle

\begin{abstract}
We consider a distributed system where multiple users access the same set of arms. The mean rewards from the arms are unknown and could be different for each user (heterogeneous). No controller is available to coordinate arm selections by the users, and if multiple users select the same arm, they collide and none of them receive any reward. For such a completely decentralized system we develop algorithms that aim to achieve the highest expected total reward in the system (optimal assignment). Due to the lack of any direct communication between the users, we allow each user to exchange information by transmitting in a specific pattern and sense such transmissions from others. However, such transmissions and sensing for information exchange do not add to network reward. For the narrowband sensing scenarios, we first develop Explore-and-Commit algorithms that converge to near-optimal allocation with high probability in a finite number of rounds. Building on this, we develop an algorithm named Explore-Signal-Exploit (ESE) that gives near logarithmic regret. We then develop a variant of ESE named ESE1 that enjoys optimal logarithmic bounds. Further, we extend the algorithm to handle the dynamic case where the players can enter and leave the system. We validate our claims through extensive experiments and show that our algorithms perform significantly better than the state-of-the-art CSM-MAB, \de, and \dets  \; algorithms.

\end{abstract}


\section{Introduction}
\label{sec:Intro}
In distributed systems, a set of players have access to a common set of actions/resources to select from, but a central controller may not always exist making the coordination among the players challenging. Further, the benefit of playing each action could be statistically different across players. Thus for effective utilization of resources, players not only need to learn the arm characteristics experienced by them but also that experienced by the others. Such setup arises in Cognitive radio networks (CRNs) where multiple players access a common set of channels in the absence of a central coordinator and mean rate each player get on a channel could be different. In this work, we develop distributed learning algorithms using the framework of multi-player multi-armed bandits. We devise signaling schemes that help the players establish coordination among themselves to achieve the common objective of maximizing the system performance.

\subsection{Multi-Player Multi-Armed Setting}
The standard Multi-Armed Bandit (MAB) setting involves a single player who aims to find the action with the highest mean reward. For the case involving multiple players, multi-player bandits are widely used where the goal is to find an assignment of players to actions such that sum reward across all players is maximized. Multi-player bandits are widely used in CRNs \cite{JSAC11_DistributedLearning_Anadakumar,TMC11_CongnitiveMediumAccess_LaiGamalJiangPoor}, where channels are cast as arms of bandits, rates/throughput received on them as rewards and the users as players. The standard MAB deals with a single player in the network where the goal is to learn a policy that minimizes the regret \cite{Book12_RegretAnalysis_Bubeck}, where regret of a given policy is defined as the difference between best expected cumulative reward (obtained with full knowledge of mean rewards) and that obtained by the given policy. In the multi-player setup, the regret is defined as the difference between the best expected cumulative network reward (sum over all players) with full knowledge of mean rewards and that obtained from the given policy. When multiple players play the same arm, we refer to the event as a collision. Collisions make the multi-players setting more challenging than the single-player setting where collisions do not occur. Also, unlike the single-player case, in multi-players bandits players will not receive a reward in each slot -- in slots where collisions are incurred or signals for coordination are sent, no reward is collected and thus adds to regret. 

We refer our setting to be \textit{static} when the players in the system do not enter or leave throughout the algorithm. In realistic scenarios, the number of players can change with time. This is referred to as \textit{dynamic} setting. We first develop an {\em explore-and-commit} algorithm that reaches near-optimal allocation quickly and then build on it to develop an algorithm that has sub-linear (logarithmic) regret. With minimal modifications, we extend our algorithms to handle the dynamic setting.

\subsection{Optimal Assignment and Performance Metric}
Achieving optimal performance i.e. maximizing sum of mean rewards obtained for all players in our setup is equivalent to solving a maximum weight-matching problem on a weighted bipartite graph (or an assignment problem). Here, the players and arms correspond to the two sets of nodes that are to be paired and the mean rewards correspond to the edge weights. If mean rewards seen by all the players are known (referred to as full information), the players can apply any max-weight matching algorithm, e.g. Hungarian method \cite{NRL1955_TheHingarianMethod_Kuhn, IAT1957_AlgorithmForTheAssignment_Munkres}, to obtain an optimal assignment and occupy the arms as per the same.

Achieving the optimal assignment in a heterogeneous setting in a finite time is an impossible task as it requires all players to not only learn their arm characterization accurately but also exchange this information with all other players without any errors. We thus focus on developing a Probably Approximate Correct (PAC) algorithm that guarantees a near-optimal assignment with high confidence. Our PAC metric for multi-player MAB is a natural generalization of PAC metric used in pure exploration setting of the standard MABs \cite{JMLR2006_ActionEliminationAndStoppingConditions_DarMannorMansour,JMLR2004_TheSampleComplexityOfExploration_MannorTsitsiklis}. This algorithm leverages the fact that if each player learns the mean rewards of arms experienced by herself and that by others within an $\epsilon > 0$ accuracy, then the optimal assignment computed by them from their estimates gives a network reward within a $2\epsilon N$ neighborhood of the optimal value (see Thm. \ref{thm:ApproxHun}). Here $N$ denotes the number of players in the network. Thus, once all players gather `sufficiently' many samples of arm rates and send `sufficiently' many signals to indicate their observations, the system can reach a near-optimal assignment. We develop efficient signaling schemes that allow players to exchange their estimates so that all of them have `near full information' of system rewards. We then develop algorithms to achieve logarithmic regret that runs in epochs of increasing sizes where each epoch includes exploration, signaling and exploitation phases. Our contributions can be summarized as follows:

\begin{itemize}


\item We develop explore-and-commit algorithms achieving near-optimal allocation with high probability in the finite rounds. 

\item We develop an algorithm named Explore-Signal-Exploit (ESE) and a variant named ESE1 that guarantees near-logarithmic and logarithmic regret, respectively under the assumption that the optimal assignment is unique. To the best of our knowledge, ESE1 is the only algorithm that guarantees logarithm regret without requiring any problem-specific information and any information on the time-horizon (any time algorithm). 

\item We consider the special case of homogeneous rewards and given a simplified variant of ESE1 named ESE2 that achieves logarithm regret with much-improved system constants. Specifically, it is of the order $O(K \log T)$, where $K$ is the number of arms.

\item With minor changes in the signaling protocol, we extend the ESE1 to handle the dynamic case. Unlike many algorithms in the literature, our algorithm does not require to restart when there is a change in the number of players (epoch-free).

\item We validate the performance of our algorithms via numerical experiments and show that the performance-gains of our algorithm are significant, compared to the state-of-the-art algorithms, and the gains improve with the increase in the number of players, $N$, in the network/system.  
\end{itemize}

\noindent
{\bf Organization of paper:} In Section \ref{sec:Model}, we describe the model and setup the performance metrics. Section \ref{sec:Perturb} bounds degradation in network reward due to lack of full knowledge of mean rewards. In Section \ref{sec:ECAlgorithm}, we consider the static case and develop an explore-and-commit algorithm that guarantees a near-optimal assignment with high confidence. In section \ref{sec:RepeatSensing} we develop Explore-Signal-Exploit (ESE) and show that expected regret achieved by its variant, named ESE1,  is at most logarithmic in $T$.
In Section \ref{sec:homo}, we describe how ESE1 can be made to improve performance when the arm gains are homogeneous. We provide a discussion on algorithm initialization in section \ref{sec:RndmHppng}. We illustrate how our algorithm scales in the dynamic scenario in section \ref{sec:dynamic}. 
Numerical experiments in Section \ref{sec:Simulation} validate our claims. Conclusions and future directions are given in Section \ref{sec:Conclu}.

\subsection{Related work}
\label{subsec:literature}
Several works have addressed the problem of learning and coordination in multi-player ad hoc networks. In this subsection, we focus on literature employing the multi-armed bandit (MAB) based approach for arm selection, as it outperforms other approaches employing random or sequential-hopping-based arm selection \cite{TSP10_DistributedLearning_LiuZhao}.

The $\rho^{rand}$ \cite{JSAC11_DistributedLearning_Anadakumar} approach employs MAB based upper-confidence bound (UCB) algorithm for arm characterization and selection. The UCB is combined with the rank-based-randomization approach to orthogonalize players in the best $N$ arms, where $N$ denotes the number of players in the network. Subsequent algorithms in \cite{WC2016_DistributedStochchasticLearning_ZhadiDongGrami,TSP14_DistributedStochastic_GaiKrishnamachari} are based on $\rho^{rand}$ and they offer further improvement in performance by reducing the number of collisions among players. However, major drawbacks of these algorithms are that they need prior knowledge of $N$ and require that all arms are identical for all players (homogeneous arms). Recently, various algorithms \cite{ICML16_MultiplayerBandits_RosenkiShamir,KDD14_ConcurrentBandits_AvnerMannor,WIOPT2013_TrekkingBased_KumarYadavDarak,TIT14_DecentralizedLearning_KalthilNayyarJain,INFOCOM16_MultiUserLax_AvnerMannor,TCNS2018_DecentralizedLearning_KalthilNayyarJain,INFOCOM2018_Low-ComplexityLearning_KangJoo}  have been proposed to improve the performance of CRNs. Among them, algorithms in \cite{ICML16_MultiplayerBandits_RosenkiShamir, KDD14_ConcurrentBandits_AvnerMannor, WIOPT2013_TrekkingBased_KumarYadavDarak} work only for homogeneous arms while algorithms in \cite{INFOCOM16_MultiUserLax_AvnerMannor, TIT14_DecentralizedLearning_KalthilNayyarJain, TCNS2018_DecentralizedLearning_KalthilNayyarJain,INFOCOM2018_Low-ComplexityLearning_KangJoo,GoT} apply to heterogeneous arms and are closer to our work.

The heterogeneous arms are considered in \cite{TIT14_DecentralizedLearning_KalthilNayyarJain, TCNS2018_DecentralizedLearning_KalthilNayyarJain, INFOCOM16_MultiUserLax_AvnerMannor, INFOCOM2018_Low-ComplexityLearning_KangJoo, GoT}. The \de and \dets\; algorithms in \cite{TIT14_DecentralizedLearning_KalthilNayyarJain,TCNS2018_DecentralizedLearning_KalthilNayyarJain} employs distributed Bertsekas auction algorithm where players in a static network agree upon an allocation via the exchange of bids. Their method finds a near-optimal allocation within a finite number of rounds. Both the algorithms assume narrowband sensing and achieve near-logarithmic regret bounds. The CSM-MAB algorithm in \cite{INFOCOM16_MultiUserLax_AvnerMannor} aims to achieve a stable allocation, assuming wideband sensing capability. CSM-MAB guarantees convergence to stable allocation in finite time with high probability. The GYRO algorithm in \cite{INFOCOM2018_Low-ComplexityLearning_KangJoo} assumes a centralized system, where players communicate their arm estimates to the controller who finds an allocation with low computational complexity.  
The authors in \cite{GoT} consider a static setting where players do not change and propose a fully-distributed algorithm with $\log^2(T)$ regret bound. Finally, the authors in \cite{kaufmann2019new} also consider a static setting with heterogeneous arms. They present a distributed Exploit-Then-Commit type algorithm with $\log(T)$ regret when the optimal allocation is unique and run-time of algorithm $T$ is known apriori (not any time algorithm).  

Our work differs from the rest of literature as we consider the more challenging setting with an unknown number of players that can enter and leave system arbitrarily and see different gains across the arms  (heterogeneous arms). We set our goal to achieve optimal allocation in the system and develop a distributed algorithm motivated by the Hungarian algorithm \cite{NRL1955_TheHingarianMethod_Kuhn}, that achieves $\log(T)$ regret bound without any information of the rewards during a collision or run-time $T$. Moreover, other works have assumed that an initialization phase gives orthogonal allocations. However, our algorithm and results are independent of this assumption.

Part of this work is presented at INFOCOM 2019 \cite{Infocom2019_DistributedLearning_TibrewalPatchalaHanawal}. The conference version gives expected regret bounds that are conditional. This extended version improves the algorithms and derives regret bounds that are unconditional. The regret analysis is significantly improved and bounds are made more explicit in all the problem specific parameters. In the process, some of the imprecise statements of the conference version made precise.
Further, the current version studies the special case of homogeneous networks and derives tighter regret bounds. Extension of the algorithms to the dynamic case only appear in this work. 

\section{Model and Setup}
\label{sec:Model}
Let $N$ and $K$ denote the number of players and number of arms, respectively. We assume $N \leq K$ and denote the set of players and arms by $[N]$ and $[K]$ respectively. The reward on each arm is stochastic and may not be the same for all the players. Let $S^n_k \coloneqq \left \{S_k^n(t)\right\}_{t\geq 1}$ denote the stochastic process that governs the reward assignment for player $n$ on arm $k$. For each $n \in [N],k \in [K]$ we assume that the process $S_k^n$ is an independent and identically distributed (I.I.D) process with a common mean $\mu_{n,k}$. The rewards are bounded with support $[0,1]$ and are independent across arms. 


A player can play or observe an arm in each round. If a player plays an arm,  she will receive a reward if she happens to be the only one to play that arm. Otherwise, she will not receive any reward. We refer to the latter event of more than two players playing the same arm as a `collision'. If a player only observes an arm, she can detect if any other player(s) is playing it, but neither receives any reward nor knows how many players playing the arm.

Let $X^n_k(t)$ denote the reward observed by player $n$ from arm $k$ in round $t$. $X^n_k(t)$ takes the following values
\begin{equation*}
X^n_k(t)=
\begin{cases}
0 \quad \mbox{if player observes}, \\
0 \quad\mbox{if  player plays and incurs collision,}\\
\mbox{$S^n_k(t)$ if player plays without collision}.
\end{cases}
\end{equation*}
We assume time is slotted. In slot $t$,  expected total reward in the system is $\mathbb{E}\left [ \sum_{n \in [N]} X^n_{a^n_t}(t)\right ] $, where $a^n_t$ denotes the arm selected by player $n$ in slot $t$. The expected reward in the system, henceforth simply referred to as system reward, is to be maximized as part of the social goal of the players. But, neither can the players talk to each other directly nor does a controller exist for coordination. The only way a player can exchange information with others is by signaling $0-1$ bits. For example, a player can signal bit $1$ or $0$ to others if she plays or does-not-play, respectively, on an arm while the others observe it.  

Let $\alpha \coloneqq (\alpha^n)_{n \in [N]}$ denote a collection of policies, where $\alpha^n \coloneqq (\alpha^n_t)_{t \geq 1}$ denotes the policy used by player $n$. The policy of each player is to select an arm and either play or observe it. The decision of each player is based only on their past observations. Let $\mathcal{H}^n_{t+1} \coloneqq \{(a_1^n,X_{a_1^n}^n,Y^n_1), \ldots, (a_t^n,X^n_{\alpha_t^n},Y^n_t)\}$ denote the history available with player $n$ in round $t+1$, where $Y^n_t$ denotes the information collected from other players through signaling. Then, the policy of player $n$ is given by a sequence of maps $\alpha_t^n: \mathcal{H}_{t-1} \rightarrow [K]$. The decision function $\alpha_t^n$ also encapsulates the information of whether to play or observe on the selected arm. However, we suppress this information to avoid clutter in notations, and abusing notation, use $\alpha_t^n$ to denote an action. We set $\mathcal{H}_0^n=\emptyset$ for all $n \in [N]$.

Let $\pi: [N] \rightarrow [K]$ denotes an assignment such that no two players are assigned the same arm. We denote set of all assignment functions as $\Pi$. If all the players know the mean reward matrix $M=\{\mu_{n,k}\}$,  then the optimal policy is to use an assignment given by
\[ \pi^* \in \arg\max_{\pi \in \Pi}\sum_{i \in [N]} \mu_{i,\pi(i)}.\] 

To find the optimal solution $\pi^*$ in a distributed fashion requires each player to sample arms prohibitively large number of times and  also require many signaling rounds to exchange information with other players, making it infeasible for all practical purposes. We thus focus on learning 'approximate' optimal solution with high probability, defined as follows:

\begin{defn}
	For a given $\epsilon>0$ and $\delta \in (0,1)$, a policy $\alpha$ is said to be $(\epsilon, \delta)$-optimal on a mean reward matrix $M$ if there exists a positive integer $T:=T(M)$ such that
	\begin{equation}
	\Pr\left \{ \sum_{n \in [N]} \mu_{n,\alpha_t^n} \geq  \sum_{n \in [N]} \mu_{n,\pi^*(n)} - \epsilon \right \} \geq 1-\delta \;\;\; \forall \;\;t \geq T.
	\end{equation}
\end{defn}

The above definition can be viewed as generalization of probably-approximately-correct (PAC) performance guarantee in the pure exploration MAB problems with single player  \cite{JMLR2006_ActionEliminationAndStoppingConditions_DarMannorMansour,JMLR2004_TheSampleComplexityOfExploration_MannorTsitsiklis} to the multi-player case.

We define expected regret of a policy $\alpha$ over a period $T$ as 

\begin{equation}
\label{eqn:Regret}
R(T, \alpha)= T \sum_{n \in [N]} \mu_{n, \pi^*(n)} - \mathbb{E}\left[ \sum_{t=1}^{T}\sum_{n \in [N]}{X^n_{ \alpha_t^n}}\right].
\end{equation}
We aim to develop policies that give $(\epsilon, \delta)$-optimal performance within a short time period $T$. Also, for a given $T$, we develop an algorithm that has sub-linear regret.

\noindent
{\bf Observation model:} Any policy needs to incorporate an appropriate signaling mechanism so that the players can exchange information with each other. In our setup, information exchange happens through observing arms. We assume that each player can observe only one arm in each round. Further, for sake of comparison, we also look at the case where multiple arms can be observed simultaneously as in \cite{INFOCOM16_MultiUserLax_AvnerMannor}. This is referred to as wide-band setting and please see the appendix for more details. The wide-band setting makes the observation model lot simpler and this helps in improved algorithm performance. We do not incur any collisions because we use observation-based model rather than the collision-based model used in \cite{kaufmann2019new} for signaling. We stress that the observation model is no weaker than the collision model in the cognitive radio setup as the player that intends to observe an arm simply sense and do not transmit on it. 

\section{Optimal Allocation with Perturbed Matrix}
\label{sec:Perturb}
Given full knowledge of the mean reward matrix $M$ to all the players, each player can find an optimal assignment using well known bipartite matching methods like Hungarian algorithm \cite{NRL1955_TheHingarianMethod_Kuhn,IAT1957_AlgorithmForTheAssignment_Munkres} and play the arm as specified by the optimal assignment. However, in our setting each player can only estimate her mean rewards and signal the same to other players. Hence the players can only have an estimate of $M$, which at best can be guaranteed to be within a small interval around the true values. An optimal assignment obtained using the estimated matrix can be sub-optimal with respect to the true reward matrix. The following result bounds the loss incurred by playing this sub-optimal assignment.
Denote by $f(M, \pi) \coloneqq \sum_{i \in [N]} \mu_{i, \pi(i)}$, the network reward by applying assignment $\pi$ on a reward matrix $M$.  

\begin{thm}
	\label{thm:ApproxHun}
	Let $M=\{\mu_{n,k}\}$ and $\hat{M}=\{\hat{\mu
	}_{n,k}\}$ be any mean reward matrices such that $|\mu_{n,k}-\hat{\mu}_{n,k}| \leq \epsilon  \ \forall \ n \in [N], k\in [K]$  for some $\epsilon >0$. Let $\pi^*$ and $\hat{\pi}^*$ denote the optimal allocations corresponding $M$ and $\hat{M}$ respectively. Then
	\[ -2N\epsilon\leq f(M, \pi^*) - f(M, \hat{\pi}^*) \leq 2N\epsilon.\]
\end{thm}
\textit{Proof}:	We prove the result from the following inequalities: 
\begin{gather}
-N\epsilon \leq f(M,\hat{\pi}^*) - f(\hat{M},\hat{\pi}^*) \leq N\epsilon \label{eqn:perturb1}\\
-N\epsilon \leq f(\hat{M},\hat{\pi}^*) -  f(M,{\pi}^*) \leq N\epsilon. \label{eqn:perturb2}
\end{gather}
The first inequality is obvious, since for any $\pi\in \Pi, f(M,\pi) - f(\hat{M},\pi)=f(M-\hat{M},\pi)$. Using the fact that each element in $\hat{M}$ is far from $M$ by at most $\epsilon$ and using $\pi = \hat{\pi}^*$ (\ref{eqn:perturb1}) follows.
\\
Note that adding or subtracting the same constant to all elements of $M$, does not change the optimal assignment.
\[ \therefore f(M, \pi^*)=f(M+\epsilon I, \pi^*)-N\epsilon=f(M-\epsilon I, \pi^*)+ N\epsilon.\]
Since each element of $\hat{M}$ is greater than the corresponding element in $M -\epsilon I$, $f(M-\epsilon I, \pi^* )\leq f(\hat{M}, \pi^* ) \leq f(\hat{M}, \hat{\pi}^*)$,
where the second inequality follows from the optimality of $\hat{\pi}^*$ on $\hat{M}$. This implies that $f(\hat{M}, \hat{\pi}^*) -f(M, \pi^*)  \ge -N\epsilon$.
Similarly, the other inequality follows as 
\[f(\hat{M}, \hat{\pi}^*)\leq f(M+\epsilon I, \hat{\pi}^*) = f(M,\hat{\pi}^* )+ N\epsilon \leq  f(M,\pi^* ) + N \epsilon.\]
The theorem statement thus follows from (\ref{eqn:perturb1}) and (\ref{eqn:perturb2}). \hfill \IEEEQED

This result suggests a method to achieve a system reward that is $\epsilon$-close to the optimal. Specifically, if all the players estimate each entry of $M$ within $\epsilon/2N$ accuracy and play an arm specified by an optimal allocation computed on the estimated matrix, then the system reward is $\epsilon$-optimal. The goal of estimating the elements of $M$ within $\epsilon/2N$ accuracy can be achieved in two steps. In the first step, each user estimates her mean rewards within $\epsilon/4N$ accuracy. In the second step, the players signal their estimated mean rewards within $\epsilon/4N$ accuracy to other players in a round-robin fashion. Then each player will have an estimated mean reward matrix that is within the $\epsilon/2N$ accuracy of the underlying true reward matrix. However, the issue with this approach is if the optimal assignment on the estimated matrix is not unique, then the optimal assignment obtained by each player could be different and multiple players may select the same arm resulting in collisions. We first assume that the optimal assignment computed on the estimated matrix by all the players is unique, which we justify to hold later in Subsection \ref{subsec:MultipleAllocation}. We next develop distributed learning algorithms that exploit these facts to obtain optimal assignments.

\section{Explore and Commit Algorithms}
\label{sec:ECAlgorithm}
In this section, we develop PAC algorithms. In these algorithms, players explore the arms initially and exchange their estimates with all users via signaling. The players then compute the optimal allocation using the estimated mean reward matrix and play an arm assigned by it thereafter. The challenges are to avoid collisions among the players and learn mean rewards as quickly as possible. Also, we need efficient signaling schemes for players to exchange information. In the following we use the convention that $\log(\cdot)$ denotes the natural logarithm and  $\log_2(\cdot)$ denotes the logarithm to base $2$.

\subsection{Algorithm}
\label{sec:Wide-band Sensing}
In this subsection, we develop an algorithm named  Distributed Optimal Assignment (DOA) for optimal assignment. The inputs to the algorithm are $K,T_s,T_r$ and $T_b$ and the algorithm guarantees an $(\epsilon, \delta)$-optimal assignment within a finite number of rounds, where $\epsilon$ and  $\delta$ are dependent on $K,T_s,T_r$ and $T_b$. The algorithm consists of three phases namely 1) Random Hopping (RH) 2) Sequential Hopping (SH) and 3) Packetized Signaling (PS). These phases run sequentially to estimate the matrix $M$. In the end, all players apply the Hungarian algorithm\footnote{The Hungarian algorithm gives an assignment that minimizes the sum reward. Since we want the sum reward to be maximized, the input given to this algorithm is the negative of the rate matrix, thus ensuring the sum reward is maximized. Henceforth, it is assumed implicitly that the negative of the rate matrix is provided to the algorithm.} and play the arm given by the optimal assignment thereafter.  The pseudo-code of DOA is given in Algorithm (\ref{alg:Algo1}) which is run by each player separately. We suppress index $n$ to avoid cluttering in the presentation.

\begin{algorithm}[!h]
	\caption{DOA} \label{alg:Algo1}
	\begin{algorithmic}[1]
		\State	Input: {$K, T_s, T_r, T_b$}
		\State	Initial Orthogonal Assignment: $(N, k,n)= RH(K, T_r)$ 
		\State	 \textbf{Exploration}: $\hat{\mu}_n=SH(K, k, T_s )$ 
		\State  \textbf{Signaling}: $\hat{M}=PS(N,\hat{\mu}_n, T_b)$
		\State Find  optimal assignment $\hat{\pi}^*$ on $\hat{M}$ 
		\State \textbf{Exploitation}: Play $\hat{\pi}^*(k)$ till end
	\end{algorithmic}
\end{algorithm}
The RH phase first ensures that all players are on different arms (orthogonalized). To achieve this in the first $T_r$ rounds, each player selects an arm uniformly at random until they experience a collision-free round. Once it happens, they continue playing that arm. We refer to an arm on which a player experiences a collision-free play as its {\em reserved arm}. $T_r$ is set such that all players experience a collision-free play with high probability. At the end of the RH phase, players also learn how many players are present and also get an index. This is achieved in $K$ rounds following the $T_r$ rounds as follows -- in the ($T_r + k$-th) round, $k \in [K]$, a player with reserved arm plays arm $k$, while all others observe it. Thus, by counting on how many arms players presence is seen after $K$ rounds (including itself), each player knows how many players are there in the system (lines $(19-20)$). If there is a player with reserved arm $1$, she gets index $1$. The player with reserved arm $k>1$ gets $n$ as its index if she observes $n-1$ plays before she plays her reserved arm (line $(20)$).       
\begin{algorithm}[!h]
	\caption*{\textbf{Phase 1:} Random Hopping (RH)}
	\begin{algorithmic}[1]
		\State Input: $K,T_r$  
		\State Initialize: Set $Lock=0$
		\For{$t=1 \dots T_{r}$}
		\If{($Lock == 1$)}
		\State Select the same arm, $\alpha_t = \alpha_{t-1}$
		\Else
		\State Randomly select an arm, $\alpha_t \sim U([K])$
		\State Set $Lock=1$ if no collision is observed
		\EndIf
		\EndFor
		\State Set $k=\alpha_t$.
		\For{$t = T_r+1 \dots T_r+K$}
		\If{($t- T_r == k$)}
		\State Play arm $(k)$
		\Else 
		\State Observe on arm $(t-T_r)$
		\EndIf
		\EndFor
		\State $A=\left \{i\in [K]: \mbox{observed a play on arm } i \neq k \right \}$
		\State Set $N = |A|+1$ and $n = |\{j \in A : j < k\}| + 1$
		\State Return $N,n$ and index of current arm $k$.
	\end{algorithmic}
\end{algorithm}

The SH phase enables all the players to estimate their mean rewards. In this phase, all the players select the arms sequentially, i.e., arm $(k+1)\mod K$ is played after playing arm $k$. Since an orthogonal allocation is maintained in each round, no collisions occur in this phase. The length of the phase is set such that the mean rewards are estimated with high accuracy. 
\begin{algorithm}[!h]
	\caption*{\textbf{Phase 2:} Sequential Hopping (SH)}
	\begin{algorithmic}[1]
		\State Input: $K, T_s, k,n$  
		\State Set $\alpha_{T_{r}+K}=k$, $r_i=0 \;\; \forall i\in [K]$
		\For{$t=T_{r}+K+1 \dots KT_{s}+K$}
		\State Play arm $\alpha_t=(\alpha_{t-1} +1 )\mod K$
		\State Observe reward $X_{\alpha_t}$ and update $r_{\alpha_t}=r_{\alpha_t} + X_{\alpha_t}$
		\EndFor	
		\State Estimate $\hat{\mu}_{n,k}=r_{k}/T_{s} \; \forall k \in [K]$
		\State Return $\hat{\mu}_n= \{\hat{\mu}_{n,k}\}$.
	\end{algorithmic}
\end{algorithm}

In the PS phase, each player signals her estimated mean rewards to others and also obtains that seen by the others through their signals. This is achieved as follows -- the PS phase consists of $NK$ frames each of $T_b$ rounds in which each player binary encodes her estimates and signals this using a sequence of bits. For example, each player can encode her estimates within $\epsilon/4N$ accuracy using $T_b = \lceil \log_2 (4N/\epsilon)\rceil$ bits. Then, each player needs $K\lceil \log_2 (4N/\epsilon)\rceil$ bits to encode her estimates of all arms. The players can then play an arm as per the bit sequence of their code -- a play corresponds to bit '$1$' and an empty play corresponds to bit '$0$'. Since players can observe other arms, within $NK\lceil \log_2 (4N/\epsilon )\rceil$ time slots, all players will receive the estimates of other players within $\epsilon/4N$ accuracy. 

During signaling, the player with index $1$ starts the signaling and then the player with index $2$ and so on. When player $i$ turn comes, it signals its estimates for all arms starting from arm $1$ to $K$. In the ${ij}^{th}$ frame, player $i$ plays arm $j$ for $T_b$ rounds as per the encoding of the estimate for arm $j$ while all players $k \neq i$ receive her encoded estimates by observing arm $j$. At the end of PS phase, each player has signaled, received and decoded the same estimated mean\footnote{This is ensured by each player using the truncated estimates in the matrix instead of the actual estimates} reward matrix of size $N\times K$. 
Each player then finds the optimal assignment on $\hat{M}$ and plays it thereafter. We state the following lemmas before providing the guarantees of DOA.

\begin{algorithm}[!h]
	\caption*{\textbf{Phase 3:} Packetized Signalling (PS)}
	\begin{algorithmic}[1]
		\State Input: $N, K, T_b, \hat{\mu}_n$
		\For{$i=1,2, \dots K$}
		\For{$j=1,2, \dots K$}
		\If{i==n}
		\State Play arm $j$ corresponding to $T_b$ bits encoding of reward $\hat{\mu}_{n}(j)$
        \Else
        \State Observe arm $j$ for $T_b$ rounds
        \EndIf
	    \State Update $\hat{\mu}_{i,j}$ decoding the observed bit pattern
		\EndFor
		\EndFor
		\State Return $\hat{M}= \left \{\widehat{\mu}_{i,j}\right \}$. \Comment{Truncate to make $\hat{M}$ of size $N\times K$}
	\end{algorithmic}
\end{algorithm}

\begin{lemma}
	\label{lma:DOARHPhase}
	Let $\delta \in (0,1)$. If RH phase is run for  $ T_{r}=\left \lceil \frac{\log(\delta/2K)}{ \log\left(1-1/4K\right)} \right \rceil $ number of rounds then all the players will be orthogonalized with probability at least $1-\delta/2$.  
\end{lemma}  

\begin{lemma}
		\label{lma:DOASHPhase}
For any given $\delta \in (0,1)$ and $\epsilon>0$ set $T_s = \frac{8N^2}{\epsilon^2}\log\left(\frac{4NK}{\delta}\right)$. Then estimated  mean rewards at the end of the SH phase  are such that $|\mu_{n,k}-\hat{\mu}_{n,k}|\leq \epsilon/4N$ for all $n\in [N], k\in [K]$ with probability at least $1 - \delta/2$.
\end{lemma}

\begin{lemma}
		\label{lma:DOAPSPhase}
	For any given $\epsilon>0$ set  $T_b = \lceil log({4N/\epsilon}) \rceil$. Let $\hat{\mu}_n$ denotes the mean reward vector that a player $n\in [N]$ signals. Then estimated  mean rewards matrix $\hat{M}=\left \{\widehat{\mu}_{i,j}\right \}$ at the end of the PS phase will be  such that $|\hat{\mu}_{n,k}-\widehat{\mu}_{n,k}|\leq \epsilon/4N$ for all $n\in [N], k\in [K]$.
\end{lemma}

Proof for Lemma \ref{lma:DOARHPhase} is similar to \cite{WIOPT2013_TrekkingBased_KumarYadavDarak}[Lemma 1]. The proof for Lemma \ref{lma:DOASHPhase} is a straightforward application of the Hoeffding's concentration inequalities \cite{hoeffding1963probability} and has been omitted.

\begin{thm}
	\label{thm:DOA}
	For a given $\epsilon>0, \delta \in (0,1)$ let $T_r, T_s$ and $T_b$ be set as in the Lemmas \ref{lma:DOARHPhase}, \ref{lma:DOASHPhase} and \ref{lma:DOAPSPhase} respectively. Then the DOA policy is $(\epsilon, \delta)$-optimal after $T_r+KT_s + NKT_b$ number of rounds for any mean reward matrix $M$. 
\end{thm}
\noindent
{\textit{Proof of Theorem (\ref{thm:DOA}):}} If (A) denotes all players orthogonalized in RH phase, (B) denotes all players estimate their mean reward within $\epsilon/4N$ accuracy and (C) denotes all players signal their estimates to others within $\epsilon/4N$ accuracy,  the estimated matrix $\hat{M}$ at each player will be such that $|\hat{M}- M|\leq \epsilon I/2N$ after $T:=T_r + KT_s+ NKT_b)$ rounds, where $I$ is an all-ones $N\times K$ matrix. Theorem \ref{thm:ApproxHun} then implies that if all players play the optimal assignment obtained from $\hat{M}$, DOA policy will be $\epsilon$-optimal for all $t > T$. Hence we have
\begin{eqnarray*}
	\lefteqn{\Pr\left \{\mbox{DOA is  $\epsilon$-optimal for}\;\; t > T\right \} \geq \Pr\{\text{A,B,C holds}\} }\\
	&=& 1-\Pr\{\text{any of A,B,C doesn't hold}\}> 1-3\delta/3= 1-\delta,
\end{eqnarray*}

where the inequality follows from Lemma (\ref{lma:DOARHPhase}), (\ref{lma:DOASHPhase}) and (\ref{lma:DOAPSPhase}). \hfill\IEEEQED

\begin{remark}
Note that mean reward corresponding to player $n$ in $\hat{M}$ is $\widehat{\mu}_n$ and not $\hat{\mu}_n$, i.e., the values that players signaled to other players in the PS phase and not what she estimated in the SH phase. This is done to ensure that all players have the same $\hat{M}$ so that they find the same optimal assignment.
\end{remark}

\subsection{Handling Multiple Optimal Allocations}
\label{subsec:MultipleAllocation}
In the design of DOA, we assumed that the optimal allocation computed from $\hat{M}$ will be unique. However, there could exist multiple optimal assignments due to which the players may compute different optimal allocations. In such a case, it may happen that two players are assigned to the same arm leading to continuous collisions. 

When multiple allocations exist, if the ties are always broken in a deterministic way, for example,  in favor of a player with the smaller index, the algorithm will always return an optimal assignment that is same across all players. This is true because, given a matrix, the Hungarian algorithm is a deterministic algorithm when provided with a deterministic method of breaking ties
(see scipy implementation \cite{Scipy_HungarianAlgorithm}). At the end of a signaling phase, each player will have the same reward matrix and hence evaluate the same assignment.

\section{Algorithm for Regret Minimization}
\label{sec:RepeatSensing}
In this section, we focus on algorithms that minimize regret. The expected regret is shown at least logarithm in $T$, i.e., lower bounded as $\Omega(\log T)$ in \cite{ALT2018_MultiplayerBandits_BessonKaufma}. We first develop an algorithm, named Explore-Signal-Exploit (ESE), that has near-logarithmic regret and give a modification that achieves the optimal regret.  

The pseudo code of ESE is given in Algorithm (\ref{alg:ESER}) which is run by each player independently. To avoid clutter in notation, we drop the player's index. The algorithm runs in epochs of growing lengths and takes as inputs $K, T_r,\{T_s(l)\},\{T_b(l)\}$ and $\{\epsilon(l)\}$, where the latter three quantities are positive sequences. 

The system is initialized with an orthogonal allocation using the RH subroutine. Next, the algorithm proceeds in epochs. Each epoch consists of exploration, signaling and exploitation phases and corresponds to one run of the DOA algorithm. Firstly, the subroutine SH is called, where players explore arms to estimate their rewards using all the samples collected so far. The updated estimates are signaled by each player to others taking turns as per their index using the PS subroutine. Every player now applies the Hungarian method on the estimated rate matrix to find the optimal assignment and plays the arm she gets, for double the time than in the previous epoch. The end of the exploitation phases marks the end of the epoch and the process is repeated.

\begin{algorithm}[!h]
	\caption{ Explore-Signal-Exploit (ESE) }
	\label{alg:ESER}
	\begin{algorithmic}[1]
		\State Input: $K, T_r, T_s(l), T_b(l), \epsilon(l)\;\forall\; l\geq 1$  
		\State Initialization: $l = 1, \hat{M} = \textbf{0}$
		\State Initial Orthogonal Allocation: $(N, k,n)= RH(K, T_r)$ 
		\While{($t < T$)}
    		\State \textbf{Exploration phase}: $\Tilde{\mu}_n=SH(K, k, T_s(l))$
    			\State $\hat{\mu}_n$ = mean rewards of arms from all samples
    		\State \textbf{Signaling}: $\hat{M}=PS(N,\hat{\mu}_n, T_b(l))$
		    \State Find  optimal assignment $\hat{\pi}^*$ on $\hat{M}$ 
		    \State \textbf{Exploitation phase}: Play $\hat{\pi}^*(n)$ for $e^{l}$ rounds
            \State $l = l+1$ ; $t = t + KT_{s}(l) + NKT_b(l) + e^l$
    		\EndWhile
	\end{algorithmic}
\end{algorithm}

\subsection{Regret Analysis}
\label{sec:RegretAnalysis}

The regret of ESE is obtained by bounding the regret from the i) Sequential Hopping, ii) Signaling and iii) Exploitation phases separately.
The following notations are helpful to state results. Let $\pi_1(M)$ and $\pi_2(M)$ denote the the set of optimal and next-best assignments on matrix $M$, i.e. $\pi_1(M) \coloneqq \{\pi : \argmax_{\pi \in \Pi}  \sum_{n \in [N]} \mu_{n,\pi(n)} \}$ and $\pi_2(M) \coloneqq \{\pi : \argmax_{\pi \in \Pi-\pi_1(M)}  \sum_{n \in [N]} \mu_{n,\pi(n)} \}$. For any $\pi_1 \in \pi_1(M)$ and $\pi_2 \in \pi_2(M)$, we define
\begin{center}
$\Delta_{max} \coloneqq \max_{\pi \in \Pi} \{ \sum_{n \in [N]} \mu_{n,\pi_{1}(n)}- \sum_{n \in [N]} \mu_{n,\pi(n)} \}$\\
$\Delta_{min} \coloneqq \{ \sum_{n \in [N]}\mu_{n,\pi_{1}(n)} - \sum_{n \in [N]} \mu_{n,\pi_{2}(n)} \}$.
\end{center}
The quantity $\Delta_{max}$ corresponds to the maximum regret that can be obtained in a round and the quantity $\Delta_{min}$ corresponds to the difference in rewards between the best and the next-best allocations. In the following, we assume that the optimal allocation is unique, i.e, $|\pi_1(M)|=1$. We use shorthand notation $R(T):=R(T, \mbox{ESE})$ to denote the regret of ESE algorithm.

\begin{thm}
    \label{thm:Regret-ESER}
    Assume the players are orthogonalized at the end of RH phase. Using packetized communication, the regret bounds for ESE algorithm are as follows
    
    i) Let a lower bound on $\Delta_{\min}$ is known, i.e., $ \Delta_{min}>\epsilon_0$ for some $\epsilon_0>0$. Set $\epsilon(l)=\epsilon_0, T_s(l) = 8N^2/\epsilon_0^2$ and $T_b(l) = \lceil \log_2 4N/\epsilon_0 \rceil$ for all $l \geq 1$, then the regret of ESE is upper bounded as
    \begin{eqnarray*}
        \label{eqn:ESERREg1}
            R(T) &\leq& \Delta_{max} KN^2 (8/\epsilon_0^2 +  \log_2 (4N/\epsilon_0)+3/N+1) \log(T) \\
            &=& O\left ((N^2K/\epsilon_0^2)\log T\right).
        \end{eqnarray*}
    
    ii) If a bound on $\Delta_{min}$ is unknown, choose $\beta \in (0,1)$ and set  $\epsilon(l)=l^{-\beta/2}, T_s(l)=16N^2/\epsilon(l)^2$ and $T_b(l)=\lceil \log_2 4N/\epsilon(l)\rceil$ for all $l\geq 1$. Then
     \begin{eqnarray*}
        \label{eqn:ESERREg2}
            R(T) &\leq& 16N^2K\Delta_{max} \log^{(1+\beta)}T\\ 
             &+&  \Delta_{\max}N^2K\log T\left \{\log_2(8N)+(\beta/2)\log_2(\log T)\right\} \\ 
             &+& \Delta_{max}e^{1/ \Delta_{\min}^{2/\beta}+1}+3NK\Delta_{max}\log T\\
             &=& O\left (N^2K\Delta_{max}\log^{(1+\beta)}(T) + e^{1/( \Delta_{\min}^{2/\beta}\log 2)}\right ). \nonumber
        \end{eqnarray*}
\end{thm}
A proof is given in the Appendix.

\subsection{Improving on regret of ESE}
\label{sec:ESERmodified}
In this subsection, we propose am improvisation to achieve logarithm regret from ESE. We refer to this algorithm  as ESE1 and its pseudo-code is given in Algorithm \ref{thm:mESER}. In ESE1 we estimate $\Delta_{min}$ at the end of $l^{th}$ epoch using $\hat{M}=\{\widehat{\mu}_{n,k}\}$, which is an estimate of $M$. Specifically, each player computes 
\[\hat{\Delta}_{min}(l) \coloneqq \left\{ \sum_{n \in [N]}\widehat{\mu}_{n,\hat{\pi}_{1}(n)} - \sum_{n \in [N]} \widehat{\mu}_{n,\hat{\pi}_{2}(n)}\right\},\] 
where $\hat{\pi}_{1}$ and $\hat{\pi}_{2}$ are computed on $\hat{M}$, which is a `good' estimate of $\Delta_{min}$ and its accuracy improves with epochs. 
The following result is key to improve performance of ESE and achieve logarithmic regret bounds. The modification to ESE is as follows: Once the condition $\hat{\Delta}_{min}(l) > \epsilon(l)$ holds for some $l$, we update $\epsilon(l+1)$ to a value less than  $\hat{\Delta}_{min}(l) - \epsilon(l)$ and maintain this for the subsequent epochs. 
\begin{lemma}
\label{lma:secondAlloc}
For any $\epsilon \geq 0$ and $\Delta_{min} > 0$, if $|M-\hat{M}| \leq  \epsilon I/2N$, we have 
\begin{equation}
-\epsilon \leq \Delta_{min} - \hat{\Delta}_{min} \leq \epsilon,
\end{equation}
where $\hat{\Delta}_{min}$ is computed from $\hat{M}$.
\end{lemma}
The condition $\Delta_{min} > 0$ is required for the above statement to holds. This corrects Lemma $4$ in \cite{Infocom2019_DistributedLearning_TibrewalPatchalaHanawal} that stated the above to hold without this condition. The Proof is given in the Appendix.
\begin{algorithm}[!h]
	\caption{ESE1}
	\label{alg:mESER}
	\begin{algorithmic}[1]
		\State Input: $K, T_r, T_s(l), T_b(l)\;\forall\; l\geq 1 $
		\State Initialization: $l = 1$, $\hat{M} = \textbf{0}$, $lock = False$
		\State Initial Orthogonal Allocation: $(N, k,n)= RH(K, T_r)$ 
		\While{($t < T$)}
    		\State \textbf{Exploration phase}: $\Tilde{\mu}_n=SH(K, k, T_s(l))$
    			\State $\hat{\mu}_n$ = mean reward from all samples
    		\State \textbf{Signaling}: $\hat{M}(l)=PS(N,\hat{\mu}_n, T_b(l))$
		    \State \textbf{Exploitation phase}: Find $\hat{\pi}^*(n)$ and play for $e^{l}$ rounds
            \State Calculate $\hat{\Delta}_{min}(l)$ from $\hat{M}(l)$
            \If{$lock == False$}
    		    \If{$\hat{\Delta}_{min}(l) > 2\epsilon(l)$}
    		    \State Set $\epsilon(l+1)=\epsilon(l)$ 
    		    and $lock =True$ 
    		    \EndIf
		    \Else {} $\epsilon(l+1) = \epsilon(l)$
		    \EndIf
		    \State $l = l+1$ ; $t = t + KT_{s}(l) + NKT_b(l) + e^l$
    		\EndWhile
	\end{algorithmic}
\end{algorithm}

\begin{thm}
\label{thm:mESER} 
Let $\beta \in (0,1)$. In the ESE1 algorithm set   $\epsilon(l)=l^{-\beta/2}, T_s(l)=16N^2/\epsilon(l)^2$ and $T_b(l)=\lceil \log_2 4N/\epsilon(l)\rceil$ for all $l\geq 1$. Then the regret of ESE1 is upper bounded as 
\begin{eqnarray*}
R(T) &\leq& 16N^2K\Delta_{max}(\Delta_{\min}^{-2/\beta}+1)^\beta\left\{\Delta_{\min}^{-2/\beta} + \log T\right \} \\ 
&  +&  N^2K \left (\log_2 8N + \frac{\beta}{2}\log_2(\Delta_{\min}^{-2/\beta}+1)\right)  \left \{ \Delta_{\min}^{-2/\beta} + \log T \right\}\\
&+& (1+\Delta_{max})e^{1/\Delta_{\min}^{2/\beta}+1}+ 3NK\Delta_{max}\log T\\
&=&  O(N^2K(\Delta_{\min}^{-2/\beta}+1)^\beta\log T).
\end{eqnarray*}
\end{thm}
{\it Proof Outline}:
First thing to notice is that there exists $l^\prime$ such that $\hat{\Delta}_{min}(l^\prime) > 2\epsilon(l^\prime)$. This is because of the decreasing nature of $\epsilon(l) = l^{-\beta/2}$ and the bounded nature of $\hat{\Delta}_{min}(l) \in [\Delta_{min}-\epsilon(l),\Delta_{min}+\epsilon(l)]$ (as shown in Lemma (\ref{lma:secondAlloc})). 
After $l^\prime$ epochs in ESE1 algorithm, we set $\epsilon(l) = \epsilon(l^\prime), T_s(l) =16N^2/\epsilon(l^\prime)^2$.
Now, $ \forall$ $l > l^\prime$, $\epsilon(l) < \hat{\Delta}_{min}(l^\prime) - \epsilon(l^\prime) \in [\Delta_{min}-2\epsilon(l^\prime),\Delta_{min}) \leq \Delta_{min} $ (from Lemma {\ref{lma:secondAlloc}}) which implies $\epsilon(l) < \Delta_{min}$ and thus from here on now, we will not be incurring any regret with probability $1-\delta(l)$. 
\noindent
Since $l^\prime$ is finite, the regret incurred in the initial $l^\prime$ epochs is constant (independent of $T$). But after $l^\prime$ epochs, the regret equations are similar to the case when $\Delta_{min}$ is known in the ESE algorithm. Consequently, we obtain logarithmic regret even when $\Delta_{min}$ is unknown. The detailed proof is given in the appendix. \hfill \IEEEQED

\begin{remark}
We assume each player has enough computational resources to calculate best and second-best assignments \cite{matsui1994algorithms} in every epoch. ESE1 has logarithmic regret bound and outperforms other algorithms in literature (see Fig. \ref{fig:ESERvdE3RegretN} in Sec. ~\ref{sec:Simulation}). More importantly, it does not require any problem specific information ($\Delta_{\min}$).
\end{remark}


\begin{remark}
If optimal allocation is not unique, ESE1 may not give logarithmic regret as the criterion to stop increasing explorations with epoch may not  satisfy. But if the number of optimal allocations are known apriori, say $p$, then the algorithm can be adjusted by calculating $\hat{\Delta}_{min}$ in every epoch as the difference in the sum rewards between the $p^{th}$ best and the $(p+1)^{th}$ best allocation. This value of $\hat{\Delta}_{min}$ will ensure the regret from ESE1 will remain logarithmic. This is because at some point set of optimal allocations will be distinguished from the sub-optimal ones and the explorations lengths will not increase.
\end{remark}

\section{Homogeneous System}
\label{sec:homo}
ESE and ESE1 algorithms consider the general heterogeneous setup where the mean rewards from arms could be different for each player. The performance of these algorithms can be significantly improved if the mean rewards from arms are same for all players, i.e., rewards are homogeneous across the players. In the homogeneous case, it is enough that one of the players learns the mean rewards of all arms and the optimal allocation is simply that the players play the non-overlapping arms from the top $N$-arms. We specialize the ESE1 algorithm to include these points where an appointed leader does the job of learning and informing the others about the optimal allocation. We refer to the new algorithm as ESE2 and its pseudo-code is given in algorithm \ref{alg:ESE2}.  
\begin{algorithm}[!h]
	\caption{ESE2 (Homogeneous) }
	\label{alg:ESE2}
	\begin{algorithmic}[1]
		\State Input: $K, T_r,\epsilon(l), T_s(l)\;\forall\; l\geq 1 $
		\State Initialization: $l = 1$, $\hat{M} = \textbf{0}$, $lock = False$
		\State Initial Orthogonal Allocation: $(N, k,n)= RH(K, T_r)$ 
		\State Player with index $n=1$ is selected as Leader
		\While{($t < T$)}
		\If{Leader ($n=1$)}
    		\State \textbf{Exploration}:  Explore each arm $T_s(l)$ times. Estimate their means ($\hat{\mu}$) and find top $N$-arms ($\hat{\pi}$). Set $\hat{\Delta}_{\min}(l)=\hat{\mu}_{N}-\hat{\mu}_{N+1}$
    		  \If{$Lock == False$}
    		    \If{$\hat{\Delta}_{min}(l) > 2\epsilon(l)$}
    		    \State Fix $\epsilon(l+1)=\epsilon(l)$ and Set $Lock =True$
    		    \EndIf
    		    \Else {} $\epsilon(l+1) = \epsilon(l)$
		    \EndIf
    		\State \textbf{Signaling}: Signal top $N$-arms and Lock indicator
		    \State \textbf{Exploitation}: Play arm $\hat{\pi}(1)$ for $e^{l}$ rounds
		    
		    \Else { non Leader with index $n>1$}
		    	\State \textbf{Exploration}: Explore each arm $T_s(l)$ rounds. 
    		\State \textbf{Signaling}: Receive top $N$-arms and Lock Indicator
		    \State \textbf{Exploitation}: Play arm $\hat{\pi}(n)$ for $e^l$ rounds
            \If{$lock == True$}
    	\State	     $\epsilon(l+1) = \epsilon(l)$
		    \EndIf
		    \State $l = l+1$ ; $t = t + KT_{s}(l) + KT_b(l) + e^l$
		   \EndIf
    \EndWhile
	\end{algorithmic}
\end{algorithm}
The modifications are as follows: 
\begin{itemize}
\item After the indexing, a leader is chosen. By default user with index, $1$, is chosen as Leader.
    \item During the exploration phase, all players explore the arms in a sequential fashion. However, only the leader estimates the mean rewards of arms and identify the top $N$-arms.
    \item During the signaling phase, the leader communicates the list of top $N$-arms while the others receive the list. The leader also signals an indicator (Lock) which indicates the other players whether the Leader has found the list of top $N$-arms with sufficient accuracy. 
    \item In the exploitation phase, each player selects and plays one of the arms in the top $N$-arms as per their index.
\end{itemize}

We next give the regret analysis of ESE2. Let the common mean rewards that a player gets on an arm $k$ be denoted as $\mu_k$. Without loss of generality, we assume that $\mu_1 \geq  \mu_2 \geq \mu_N > \mu_{N+1} \cdots \geq \mu_K$. The requirement that $\mu_N > \mu_{N+1}$ ensures that the optimal assignment is unique.
The $\Delta_{\min}$ for this case has a simple structure as given in the following lemma. 
    \begin{lemma}
    \label{lma:deltamin_homo}
    (i) For the homogeneous case, $\Delta_{min} = \mu_{N} - \mu_{N+1}$, i.e., gap between the mean of $N$-th and $(N+1)$-th arm. 
    
    ii) If $\hat{\mu}$ estimate is such that $|\mu-\hat{\mu}| \leq  \epsilon I/2$, then 
\begin{equation}
-\epsilon \leq \Delta_{min} - \hat{\Delta}_{min} \leq \epsilon,
\end{equation}
where $\hat{\Delta}_{min}=\hat{\mu}_N- \hat{\mu}_{N+1}$ is computed from $\hat{\mu}$ after arranging its components in decreasing order.
    \end{lemma}
    \textit{Proof for part (i)}: 
    $\Delta_{min}$ denotes the difference between total reward of the best and the second-best assignment. The best assignment gives $\sum_{i=1}^N\mu_i$, and the second-best assignment gives $\sum_{i=1}^{N-1}\mu_i + \mu_{N+1}$ that can be obtained from the top ($N-1$) arms and (${N+1}$)-{th} arm. Hence $\Delta_{\min}=\mu_N- \mu_{N+1}$.\\  
   \textit{Proof for part (ii)}: With the above definition of $\Delta_{min}$, we can now write $|\Delta_{min} - \hat{\Delta}_{min}|$ as follows:
   \begin{eqnarray*}
   |\Delta_{min} - \hat{\Delta}_{min}| &=& |(\mu_N- \mu_{N+1}) - (\hat{\mu}_N- \hat{\mu}_{N+1})|\\
   &=& |(\mu_N- \hat{\mu}_N) - (\mu_{N+1} - \hat{\mu}_{N+1})| \\
   &\leq & |(\mu_N- \hat{\mu}_N)| + |(\mu_{N+1} - \hat{\mu}_{N+1})| \\
   &\leq & \epsilon/2 + \epsilon/2 = \epsilon 
   \end{eqnarray*}
   The second-last inequality follows from triangle inequality while the last inequality is a given.\hfill \IEEEQED


\begin{thm}
\label{thm:mESER_homo} 
Let $\beta \in (0,1)$. In the ESE2 algorithm set   $\epsilon(l)=l^{-\beta/2}, T_s(l)=4/\epsilon(l)^2$ and $T_b(l)=N\lceil \log_2 K \rceil +1$ for all $l\geq 1$. Then the regret of ESE2 is upper bounded as 
\begin{eqnarray}
R(T) &\leq& 4\Delta_{max}K (\Delta_{\min}^{-2/\beta}+1)^{\beta}\left\{\Delta_{min}^{-2/\beta}+ \log(T)\right\} \nonumber\\
&+ &   (N \log_2 K+N+1)\log T+  \Delta_{\max}e^{1/\Delta_{\min}^{2/\beta}+1} \nonumber\\ 
&+& 2K \Delta_{\max}\log T \nonumber\\
&=&  O(K\log (T)). \nonumber
\end{eqnarray}

\end{thm}
{\it Proof Outline}:
    In the heterogeneous setting, the best and the second-best allocations could differ at all the $N$ positions. On the other hand, in the homogeneous setting, from Lemma \ref{lma:deltamin_homo} they would differ at only one position. This eliminates a factor of $N$ in the precision-requirements in the exploration phase and hence $N^2$ in $T_s$ (from the Hoeffding's inequality). Further, due to the decrease in communication overhead, the $N^2$ term in the signaling phase disappears. The remaining proof follows similarly to the proof of Thm \ref{thm:mESER} and details are given in the appendix.
    \hfill \IEEEQED

    The signaling overhead in the homogeneous case is significantly low compared to the heterogeneous case as only the leader has to learn and signal the list of top $N$-arms. This signaling can be done using at most $N\lceil \log_2K \rceil$ as the index of at most $N$ arms is to be signaled. An additional bit is also required to indicate whether the leader has the arms list with required accuracy (Lock Indicator). This improves the regret bound to $O(K\log(T))$ as compared to the heterogeneous case with regret bound of $O(N^2K\log(T))$ as shown in the Thm.~\ref{thm:mESER_homo}. 
    SIC-MMAB works only for the homogeneous case and its regret is also upper bounded by $O(K\log(T))$. However, the analysis approach in  SIC-MMAB is based on elimination method and is different from ours.

\section{Orthogonalization in Random Hopping}
\label{sec:RndmHppng}
The results in Thms. \ref{thm:Regret-ESER}, \ref{thm:mESER}, \ref{thm:mESER_homo} are conditioned upon the event that the Random Hopping phase giving us an orthogonal allocation. But with a non-zero probability, the allocation at the end of the phase can be non-orthogonal. This section eliminates this conditioning and discusses modification to algorithms so that bounds are unconditional, i.e., the bounds take into account the possibility the allocation at the end of the phase is non-orthogonal.

\subsection{Time T is known}
Upon running the random hopping phase for a duration of  $T_{r}=\left \lceil \frac{\log(\delta_R/K)}{ \log\left(1-1/4K\right)} \right \rceil$  (from Lemma \ref{lma:RHPhase}) at the beginning of the ESE1 algorithm, the initial allocation is orthogonal with probability $1 - \delta_R$,  and with probability $\delta_R$, the system is in non-orthogonal in the subsequent round incurring collisions. Then the expected cumulative regret for $T$ rounds of the ESE1 algorithm after $T_r$ rounds of random hopping phase is as follows:
\begin{eqnarray}
\label{eqn:RegretRandomHopping}
R(T) & \leq & (1- \delta_R) R_{ESE1}(T) + \delta_R N T + N T_r
\end{eqnarray}
where the first term is the regret incurred in the ESE1 algorithm weighted by the probability of a successful Random Hopping phase, the second term is the maximum regret that can be incurred when the allocation is not orthogonal weighted by the probability of an unsuccessful random hopping phase and the third term is the maximum regret incurred during the Random Hopping Phase.

If the total run-time of the algorithm is known apriori to be $T$, then setting $\delta_R = 1/T$ would give $T_r \propto \log(T)$ and reduce Eqn. \ref{eqn:RegretRandomHopping} as follows:
\begin{eqnarray*}
\label{eqn:}
R(T) & \leq & (1- \delta_R) R_{ESE1}(T) + \delta_R N T + N T_r\\
    &  = & (1- 1/T)R_{ESE1}(T) + \frac{NT}{T} + N T_r = O(N^2K\log(T))
\end{eqnarray*}
which is of the same order as that in Thm. \ref{thm:mESER}.

\subsection{Time T is unknown}
We look at the case where we do not know the total run-time of the algorithm  apriori and come up with a modification to the random hopping phase. The modification to the ESE1 algorithm to incorporate this is to have a constant duration of random hopping phase in every epoch. 

\begin{lemma}
Let RH run in every epoch of ESE1 for a constant duration of $-\log^{-1}(1-1/4K)$. This will result in $O(N^2K(\Delta_{\min}^{-2/\beta}+1)^\beta\log T)$ regret for all without having knowledge of the run-time of $T$. 
\end{lemma}
{\it Proof:} Let $t = -\log^{-1}(1-1/4K) > 0$. Then at the end of the $l^{th}$ epoch, a total of $lt$ rounds of random hopping would have taken place, meaning that the allocation is not orthogonal with probability atmost $\delta_R(l)=Ke^{-l} $ (from Lemma \ref{lma:RHPhase}). Then the expected cumulative regret incurred in the $l^{th}$ epoch of ESE1 is as follows
\begin{eqnarray*}
    R(l) & \leq & (1-\delta_R(l)) R_{ESE1}(l) + N\delta_R(l)2^{l} + Nt \\
     &\leq &  R_{ESE1}(l) + Nt + NKe^{-l}e^{l}\\ 
    R(T) & = & \sum_{l=1}^{l=l_0}R(l)\leq  \sum_{l=1}^{l=l_0}R_{ESE1}(l) + Ntl_0 + NKl_0 \\
    &=& R_{ESE1}(T)+ Nt \log(T) + NK\log(T)\\
    & = & O(N^2K(\Delta_{\min}^{-2/\beta}+1)^\beta\log T).
\end{eqnarray*}

\section{Dynamic ESE1}
\label{sec:dynamic}
In real systems, players may join and leave the systems continuously. 
In this section, we propose an approach to extend the ESE1 algorithm to such dynamic scenarios. 

The analysis we present holds in general, but for the ease of understanding, we assume users arrive or depart uniformly at random, one at a time. But as time increases, the duration of the exploitation phase increases exponentially. Thus the probability of arrivals or departures happening in an exploitation phase is significantly more than that happening in other phases. Therefore we consider arrivals and departures happening only in the exploitation phase.

We proceed to show how arrivals and departures of users would perturb the system and how the algorithm could be modified to adapt to the situations.

\subsection{Departures}

As a protocol, when a player wants to leave the system, they must collide with every other player to signal them about their departure. Since the allocation is common knowledge in every epoch, all the players can scan through the arms to find out which user left the system. This will ensure every remaining player uniquely knows which player left the system. They can thus recompute the optimal allocation with the lesser number of players and take over the new arm as per the new allocation. 

\begin{lemma}
The ESE1 algorithm will adapt to departures and converge to optimal assignment with O($\log(T)$) regret.
\end{lemma}
\textit{Proof}: Since all current players know when a subset of players leave the network, each of them re-index themselves and calculates $\hat{\Delta}_{min}$ after the departures. Basing on the condition between $\epsilon(l)$ and $\hat{\Delta}_{min}$ from line $(10)$ in Algorithm \ref{alg:mESER}, the algorithm proceeds and adjusts to this perturbation. The system might require additional epochs to settle to the optimal assignment, but the order of regret is still $O(\log(T))$.

\subsection{Arrivals}
In case of an arrival of a player, we modify the ESE1 algorithm to accommodate the learning phase for the new player. When a new player joins, it signals to the existing players of its arrival by colliding with all other players. This is followed by scanning of all the arms by all the players to find out which arm the new player has been allocated. Once this has been signaled, the existing players and the new player start selecting the arms sequentially as per the arm index and allow the new player to learn her rewards on all the arms (exploration phase). When the players select the arms sequentially no collisions occur in the network. To ensure everyone in the system has estimates of the rates with the same precision (same $\epsilon$), player with index `1' signals to the new player the number of exploration rounds required and the number of epochs covered so far. This signaling requires only a constant number of rounds.

The new player now explores for the number of rounds specified, during which the remaining old players stay silent and incur full regret (in the worst case scenario). We could have them exploiting other arms, but for simplicity, assume otherwise. Once the new player finishes the required duration of exploration, everyone in the system now has the estimates within the same precision. They all undergo the signalling phase in which the new rate matrix is communicated, followed by exploitation and then normal execution of the algorithm.

\begin{lemma}
The above-stated modification to ESE1 algorithm will adapt to arrival and converge to the optimal assignment with $O(\log(T))$ regret.
\end{lemma}
{\it Proof:} When the new player joins, the additional exploration regret which the system incurs is proportional to the total time required for the new player to learn her rewards with the same precision. Since the player arrives at a finite time instant, the amount of time for which he would be required to do exploration, and thus the amount of time for which the system incurs full regret, would remain finite. This regret also does not scale with the total time $T$ of the algorithm. There is a regret due to signaling the new player the number of rounds to explore to achieve the same precision as the system, this is a constant independent of $T$. 

Thus in the overall regret for the system, the addition of a new player would simply add a constant regret which is dependent on the arrival instant of the player and does not scale with the total time $T$ of the algorithm. Thus the regret remains $O(\log(T))$ with the above-mentioned modifications to the ESE1 algorithm.

\subsection{Constraints on the effective arrival rate}

As seen in the previous two subsections, a single arrival or departure, results in no change in the order of the regret in the ESE1 algorithm. But if the number of arrivals occurs at certain rates, then the regret from each of the arrivals, despite being a constant, could add up to be more in order than that of ESE1. Specifically, if the effective arrivals (taking the departures into account) is lesser than $O(\log(T))$ till round $T$, then the overall order of the regret incurred in the dynamic scenario still remains $O(\log(T))$. This is summarized in the following theorem.

\begin{thm}
\label{thm:ArrivalRate}
Let the total number of arrival and departure over a period $T$ is at most $O(\log(T))$. Then the regret of ESE1 algorithm adapted to the dynamic case is at most  $O(\log(T))$.
\end{thm}

\section{Experimental Results and Observations}
\label{sec:Simulation}

To demonstrate the effectiveness of our algorithms, we present simulations and comparisons with state-of-the-art algorithms in the literature.
For all simulations, we consider that each player draws Bernoulli rewards from each arm. For each player-arm pair, the mean reward is set uniformly at random from $[0, 1]$. Our setup can be viewed as communication over a binary symmetric channel with unknown transmission rates. All performance curves are obtained by averaging $50$ simulation runs. We assumed packetized signaling in all the runs. For a fair comparison, we compare our algorithms against algorithms that use similar sensing capabilities. 
\begin{figure*}[!tb]
\vspace{-.8cm}
	\centering
	\subfloat[]{\includegraphics[scale=0.35]{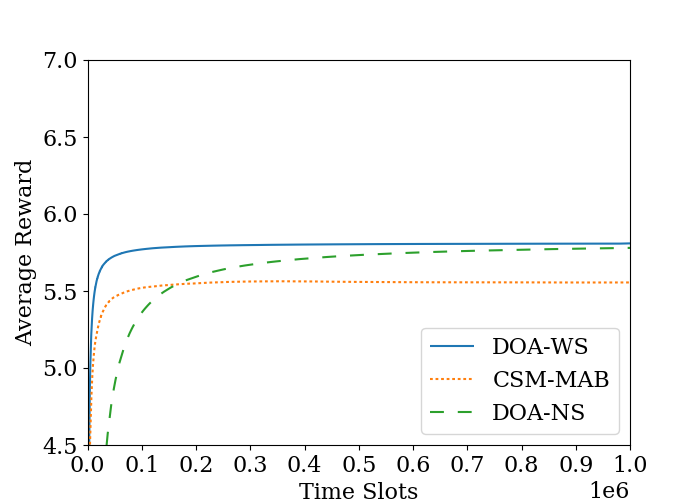}%
		\label{fig:1}}
	\subfloat[]{\includegraphics[scale=0.35]{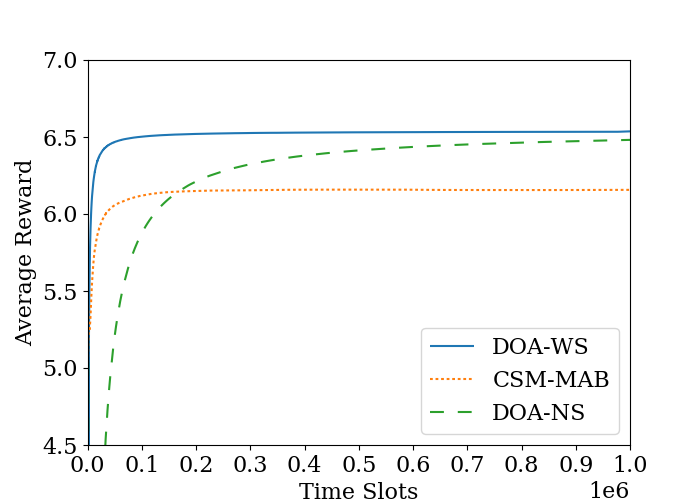}%
		\label{fig:2}}
	\subfloat[]{\includegraphics[scale=0.35]{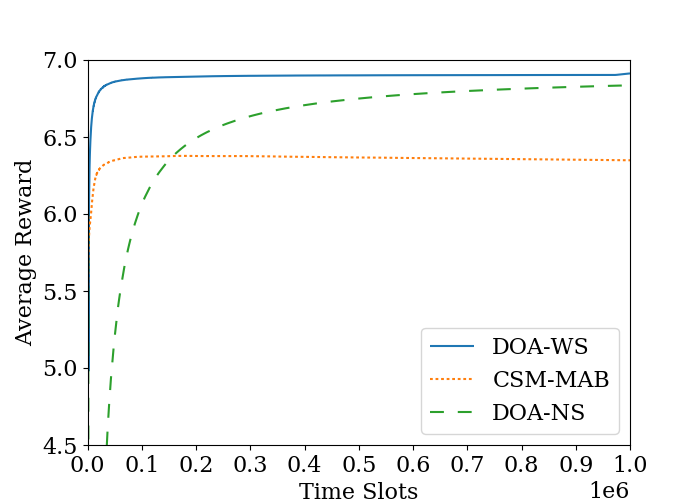}
		\label{fig:3}}
	\caption { Average reward comparison. We set $K=12$ and $N=\{8,10,12\}$ in plots (a),(b),(c) respectively.}
	\label{fig:CSMvsDOA}
\end{figure*}
\begin{figure*}[!tb]
	\centering
	\subfloat[]{\includegraphics[scale=0.35]{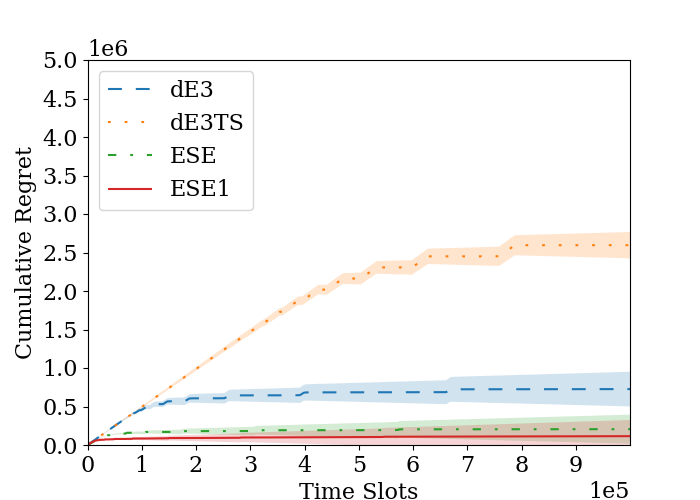}%
		\label{fig:4}}
	\subfloat[]{\includegraphics[scale=0.35]{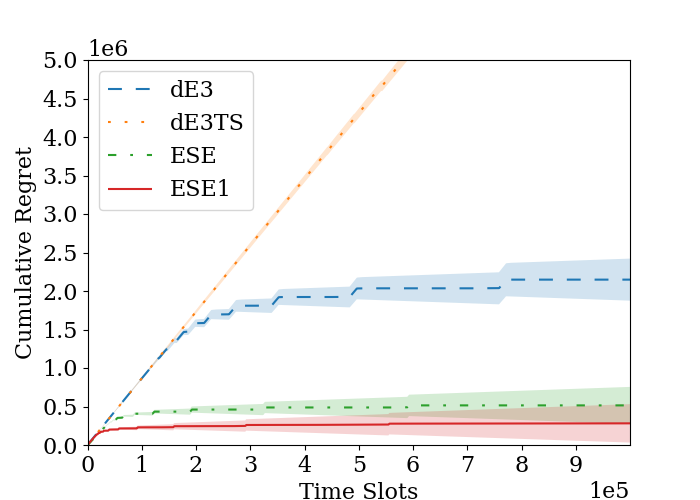}%
		\label{fig:5}}
	\subfloat[]{\includegraphics[scale=0.35]{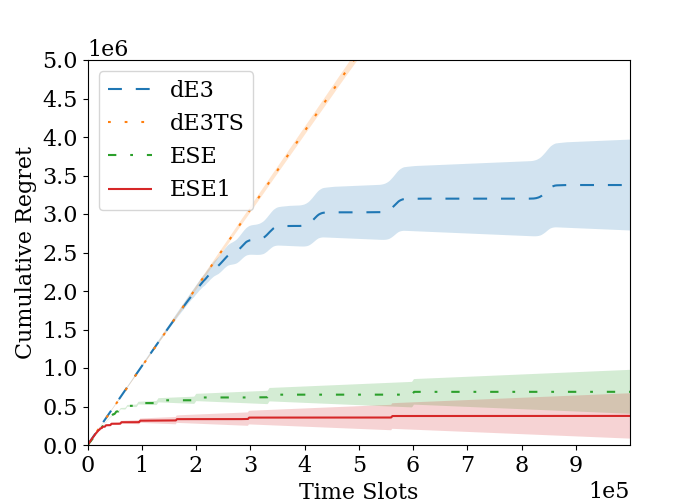}%
		\label{fig
		;6}}
	\caption {Regret Comparison (with $95\%$ confidence intervals). We set $K=12$ and $N=\{6,10,12$\} in (a), (b), (c), respectively.}
	\label{fig:ESERvdE3RegretN}
\end{figure*}



\subsection{Reward comparison in Wideband sensing setting} 
We compare the performance of DOA-WS with the CSM-MAB algorithm \cite{INFOCOM16_MultiUserLax_AvnerMannor} that also assumes wideband sensing. CSM-MAB aims to achieve stable allocation in the network by allowing players to swap channels with each other. In CSM-MAB, time slots are divided into frames of size $2K$. In each frame, an initiator (a player selected randomly) requests other players occupying channels better than her current channel, to swap channels with her. The requests are sent as per a preference list of the initiator. The frame structure is such that the initiator proposes a swap by signaling on the channel she likes to occupy, and the player occupying that channel signals back if she accepts the swap. CSM-MAB guarantees convergence to a stable allocation in finite time with high probability.

We run DOA-WS and CSM-MAB for $T=10^6$ rounds on the same problem-instance each time.  Average network reward obtained are shown in Fig (\ref{fig:CSMvsDOA}) where we fix number of channels as $K=12$ and vary the number of users as $N=\{8,10,12\}$. Similar observations were made when we fixed the number of users and varied the number of channels, but we skip those figures due to space constraints. The DOA-WS performs better than the CSM-MAB in all cases, and the improvement is significant, particularly for higher values of $N$. The performance of DOA-NS, which permits sensing of only one channel at a time, is also shown in Fig \ref{fig:CSMvsDOA}. Even though DOA-NS has restricted capabilities, it performs better than CSM-MAB eventually. The better performance of DOA-WS and DOA-NS can be attributed to the following observations:

{\it Optimal assignment v/s Stable assignment:} 
Optimal allocation is also a stable allocation. Hence  DOA-WS and DOA-NS aim to reach the `best' stable allocation among all possible allocations. In CSM-MAB, players use UCB indices to rank channels and always try to move to their best channels. It may happen that the players reach some stable allocation that is far from the optimal in the initial stages and after that no player can switch to her better channels, resulting in lower network throughput.



{\it Faster convergence:} 
A player can move to better channel in CSM-MAB only when she is the initiator. Since the initiator is chosen in a probabilistic fashion, multiple time slots pass where allocations will not improve, keeping the network in a sub-optimal `state' for a longer duration. Also, even after reaching a stable allocation, there can be collisions in the network as some players that become initiators propose swaps in each frame. However, this never happens in DOA-WS (or NS) as no collisions occur after the random hopping phase. 
 
{\it Higher number of empty channels:}
With the increase in the number of empty channels, we see in Fig \ref{fig:CSMvsDOA} that the performance gain DOA-WS has over CSM-MAB reduces. Intuitively, this is because the initiators will have higher degrees of freedom with more empty channels thus increasing their chances of reaching a better stable allocation. 
\subsection{Regret Comparison in Narrowband Sensing}

We compare the performance of the ESE and ESE1 algorithms against \de and \dets \; algorithms given in \cite{TCNS2018_DecentralizedLearning_KalthilNayyarJain}. All four algorithms assume a narrowband sensing scenario. 

\de and \dets \;  alternate between exploration and exploitation phases with an auction mechanism in between the phases. In the exploration phase, each player samples the channels for a fixed number of rounds ($\gamma$) in a round-robin fashion (only one player explores at a time). After every exploration phase, players use Bertsekas auction mechanism to find an $\epsilon$-optimal allocation. Players then exploit the allocation for the exponentially large number of rounds (growing over phases). We set $\gamma=100$ and $\gamma=400$ for \de and \dets \;   respectively, and $\epsilon=0.001$ as suggested in \cite{TCNS2018_DecentralizedLearning_KalthilNayyarJain}[Sec. V].  We set $T_s = 100$ in ESE and ESE1. We use packetized signaling for all four algorithms and the number of bits is set according to $\epsilon$ accuracy. All the algorithms are run for $T=10^6$ rounds. 

Fig \ref{fig:ESERvdE3RegretN} demonstrates cumulative regret plots of \de, \dets \;, ESE and ESE1 for a fixed number of channels $K=12$ and varying number of users $N=\{6,10,12\}$. As seen, ESE1 and ESE perform significantly better than both  \de and \dets. Also, from Thm.~\ref{thm:mESER}, we know that ESE1 achieves logarithmic regret while ESE (Thm. \ref{thm:Regret-ESER}), \de and \dets\;  \cite{TCNS2018_DecentralizedLearning_KalthilNayyarJain} all incur near-logarithmic regret. This performance improvement is also visible in Fig. \ref{fig:ESERvdE3RegretN}.    
When $N$ increases, the performance of our algorithms improve significantly which can be attributed to the following two reasons:
%
%
%

{\it Simultaneous exploration:} Recall that in the SH phase all players are on different channels and sample arms simultaneously without colliding with each other. Whereas in both \de and \dets \;, the players take turns to explore the arms which makes learning slow.

{\it Bipartite Matching}
The number of iterations required by the Bertsekas auction algorithm is of order $O(N^2/\epsilon)$ and hence increases fast with the number of players. Further, as the number of players increases, the number of "losing bids" increases. Every player corresponding to a "losing bid" does not have any allocation at the end of the auction iteration and thus incurs heavy regret for that round. Thus, for a $N$ user system, the time complexity for communication through auctioning is $O(N^3/\epsilon)$ whereas our algorithms perform signaling in $O(N^2)$.

\begin{figure}[!b]
\vspace{-.8cm}
\includegraphics[scale=0.3]{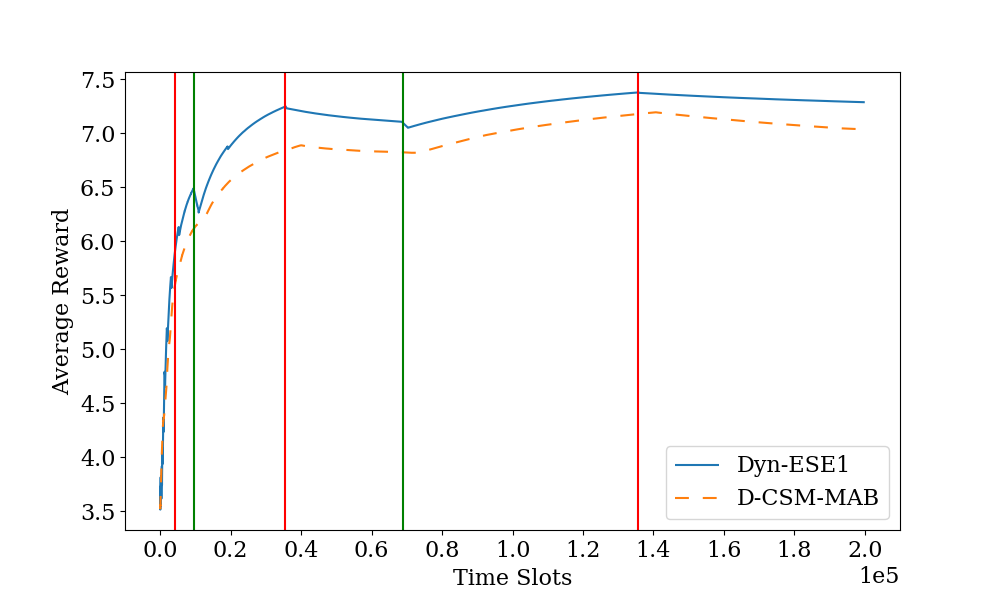}%
		\centering
	\caption {Average reward comparison in dynamic case. We set $K=10$ and $N$ varies between 7 to 9. \label{fig:dynamic}}
\end{figure}

\subsection{Reward comparison in the Dynamic Setting}
We look at the performance of the ESE1 algorithm in the dynamic setting, where the number of users changes with time. We compare the average reward of our algorithm with D-CSM-MAB algorithm of \cite{avner2018multi}, which also considers users varying over time. 
We run both the algorithms for $2*10^5$ rounds for $K=10$ channels and a varying number of users. As shown in Fig. \ref{fig:dynamic}, the arrivals, and departures are indicated as green and red vertical lines respectively. 
We expect that during departures, our algorithm immediately adjusts to this perturbation and can recalculate optimal system reward. This is observed near the red vertical lines in the figure. Further, we observe that our algorithm incurs some regret during arrivals but quickly converges to the optimal allocation and maintains a higher average reward compared to the D-CSM-MAB. This transition time comes from the initial exploration performed by the newly arrived user in the system. Since this time is of the order-logarithmic in the time elapsed, it ensures that the regret remains of order - $\log(T)$.

\section{Conclusion}
\label{sec:Conclu}
We studied an ad hoc network with multiple players with each player experiencing different and unknown rates on different channels. There is no central coordinator to facilitate communication between the players and the goal is to achieve the highest possible throughput in the network in a distributed fashion. We provide explore-and-commit algorithms that guarantee total reward and converges close to optimal with high confidence in finite time. We then extended the algorithm to achieve logarithmic regret, even in cases where the number of users can change with time. To the best of our knowledge, ours is the only algorithm that guarantees logarithmic regret in the multi-player setting without requiring to know any problem-specific details (sub-optimality gap) and the time horizon $T$. We showed that our algorithms perform significantly better than the state-of-the-art CSM-MAB, \de, and \dets \; algorithms.

In our work, we primarily focused on distributed algorithms. It is interesting to establish the lower bounds and to know what is the best one can achieve. Though our algorithm ESE1 has a $\log (T)$ dependence on $T$, its dependence on $N$ is $O(N^2)$. It is interesting to see if this can be improved. 



\section*{Appendix}
\label{sec:proofs}
\subsection{{\bf Proof of Lemma \ref{lma:secondAlloc}}}
Let $\hat{M}(l)$ be the estimated matrix after learning and signaling in the $l^{th}$ epoch such that $|M-\hat{M}| \leq \epsilon I$. Let $\hat{\pi}_1(l)$ and $\hat{\pi}_2(l)$ be the best and second-best allocations on $\hat{M}(l)$ respectively as calculated in ~\ref{sec:RegretAnalysis}. Let $\pi_1$ and $\pi_2$ be similar quantities in case of the matrix $M$.
Also, we define $\hat{\Delta}_{min}(l) = f(\hat{M}(l),\hat{\pi}_1(l)) - f(\hat{M}(l),\hat{\pi}_2(l))$.

\textit{Case 1}: If $\pi_1=\hat{\pi}_1$, $f(\hat{M},\hat{\pi}_2) \leq f(M+\epsilon I,\hat{\pi}_2) \leq f(M+\epsilon I,\pi_2)$
where the second inequality holds because $\pi_1=\hat{\pi}_1 \neq \hat{\pi}_2$ and, due to uniqueness of $\pi_1$, $\hat{\pi}_2$ cannot give better reward on $M$ than $\pi_2$. 

\textit{Case 2}: If $\pi_1 \neq \hat{\pi}_1$,
$f(\hat{M},\hat{\pi}_2) \leq f(\hat{M},\hat{\pi}_1) \leq f(M+I\epsilon,\hat{\pi}_1) \leq f(M+I\epsilon,\pi_2)$ where the last inequality holds because $\pi_1 \neq \hat{\pi}_1$ and thus $\hat{\pi}_1$ cannot give better reward than $\pi_2$ on $M$. \\
$ \because f(M+\epsilon I,\pi_2) = f(M,\pi_2)+N\epsilon$, $f(\hat{M},\hat{\pi}_2) - f(M,\pi_2) \leq N\epsilon$.

A similar comparison of $\hat{M}$ with $M-\epsilon I$ instead of $M + \epsilon I$ proves that $f(M,\pi_2) - f(\hat{M},\hat{\pi}_2) \leq N\epsilon $. 
\noindent
Now using the relations in equations (\ref{eqn:perturb1}) and (\ref{eqn:perturb2}) we get $-2N\epsilon \leq \Delta_{min} - \hat{\Delta}_{min} \leq 2N\epsilon$ Replacing $\epsilon$ by $\epsilon/2N$ completes the proof. \hfill \IEEEQED

\subsection{{\bf Proof of Theorem \ref{thm:Regret-ESER}}}
In the following, we denote the natural logarithm as $\log (\cdot)$ and that with  base $2$ as $\log_2(\cdot)$. 

\noindent
{\it Proof of i)}:
Let $\epsilon_0$ is such that $\Delta_{min}>\epsilon_0$. Set $T_s:=T_s(l) = 8N^2/\epsilon_0^2$ and $T_b =T_b(l)=\lceil log_2(4N/\epsilon_0) \rceil$ for all $l$. For a given $T$, let $l_0$ denote the phase in the algorithm terminates. We have $T\geq \sum_{l=1}^{l_0}(T_s(l) + T_b(l) + e^{l})\geq e^{l_0}$ and $\log(T) \geq l_0$. 
The regret from the exploration phase is bounded as
\begin{equation*}
\begin{array}{lcl}
R^O(T) 
&\leq & \sum_{l=1}^{l=l_0} \Delta_{max} T_s K
 \leq  \Delta_{max} K T_s log(T)
\end{array}
\end{equation*}
Regret from the signalling phase is bounded as 
\begin{equation*}
\begin{array}{lcl}
R^{S}(T) 
& \leq &  \sum_{l=1}^{l=l_0}\Delta_{max}T_b KN^2 \leq \Delta_{max}  N^2K T_b\log(T) \\
\end{array}
\end{equation*}
Since $\epsilon_0 < \Delta_{min}$, the expected regret in $l$-th epoch would be bound as $R^E(l) \leq e^{l}\Delta_{max}\delta(l)$, where  $\delta(l)$ is the probability that allocation is not optimal. Summing over all epochs, the expected regret in the exploitation phase is given as
\begin{equation*}
    \begin{array}{lcl}
R^E(T)
    &= & \Delta_{max}\sum^{l_0}_{l=1}e^lPr\{ |\hat{M}(l)-M|>\epsilon_0/2N  \}\\
    & \leq & \Delta_{max}\sum^{l_0}_{l=1}e^{l}3NKe^{-\epsilon_0^2lT_s/8N^2}\\
    & \leq & 3NK\Delta_{max}\sum^{l_0}_{l=1}(ee^{-\epsilon_0^2T_s/8N^2})^{l}\\
    & \leq & 3NK\Delta_{max}\sum^{l_0}_{l=1}(ee^{-1})^{l} \leq 3NK\Delta_{max}\log T.
\end{array}
\end{equation*}
{\it Proof of part ii)}:
When a lower bound on $\Delta_{\min}$ is unknown, we adaptively set the values of sequences $\{\epsilon(l)\}, \{T_s(l)\}, \{T_b(l)\}$ as $\epsilon(l)=l^{-\beta/2}, T_s(l)=16N^2/\epsilon(l)^2=16N^2l^\beta$ and $T_b(l)=\lceil \log_2 4N/\epsilon(l) \rceil=\lceil \log_2 4N l^{\beta/2} \rceil$ for all $l$. Let $l_0$ denote the index of the last phase when the algorithm terminates in $T$ rounds. We have $\log T \geq l_0$.

\noindent
The regret from the exploration phases is given by
\begin{equation*}
\begin{array}{lcl}
R^O(T) 
&\leq &  \Delta_{max} \sum_{l=1}^{l=l_0}T_s(l)K\\
& \leq & K\Delta_{max} \sum_{l=1}^{l=l_0}T_s(l_0) \\
& \leq & 16N^2K\Delta_{max} l_0 l_0^{\beta}\leq 
16N^2K\Delta_{max} \log^{(1+\beta)}T
\end{array}
\end{equation*}

For the packetized signaling phases, $T_b(l) = \lceil log_2(4N/\epsilon(l)) \rceil\leq log_{2}(4N) + (\beta/2)\log_2(l) + 1$. Regret in this phase is  bounded as
\begin{equation*}
\begin{array}{lcl}
R^{S}(T) 
& \leq &  \Delta_{\max}\sum_{l=1}^{l=l_0}T_b(l) KN^2\leq  \Delta_{\max} N^2K \sum_{l=1}^{l=l_0}T_b(l_0) \\
&  \leq & \Delta_{\max}N^2Kl_0\left \{ \log_2(8N) + (\beta/2)\log_2 l_0 \right \}\\
&  \leq & \Delta_{\max}N^2K\log T\left \{\log_2(8N)+(\beta/2)\log_2(\log T)\right\}. \\
\end{array}
\end{equation*}
We now look at the regret from exploitation phases. Over the $l^{th}$ phase, the expected regret is bounded as $R^E(l) \leq e^{l}\left\{\epsilon(l)(1-\delta(l)) + \Delta_{max}\delta(l)\right\}$. Accumulating over all epochs, we have
\begin{equation*}
    \begin{array}{lcl}
 R^{E}(T) 
    & = & \sum^{l_0}_{l=1}e^{l}\{\epsilon(l)(1-\delta(l))+\delta(l)\Delta_{max}\} \\
    & = & \sum^{l'}_{l=1}e^{l}\{\epsilon(l)(1-\delta(l))+\delta(l)\Delta_{max}\} \\ 
    && \hspace{50pt}+ \sum^{l_0}_{l=l'+1}e^{l}\delta(l)\Delta_{max}\\
    & = & R^E_1(T) + R^E_2(T),
    \end{array}
\end{equation*} where 
$l^\prime$ is the smallest integer such that $\epsilon(l^\prime) \leq \Delta_{\min}$, i.e., $l^\prime = \lceil\Delta_{\min}^{-2/\beta}\rceil$  and 
\begin{align*}
 R^E_1(T) &=  \sum^{l^\prime-1}_{l=1}e^{l}\left\{\epsilon(l)+\delta(l)(\Delta_{max}-\epsilon(l))\right\}  \\ 
 R^E_2(T)&=\sum^{l_0}_{l=l^\prime}e^{l}\left\{\epsilon(l)+\delta(l)(\Delta_{max}-\epsilon(l))\right\}  
\end{align*}
 We next bound each term separately.
 \begin{align*}
 R^E_1(T)
    & =  \sum^{l^\prime-1}_{l=1}e^{l}(\epsilon(l)+\delta(l)(\Delta_{max}-\epsilon(l)))\\
    & \leq  e^{l^\prime}(\Delta_{max}) =\Delta_{max}e^{1/ \Delta_{\min}^{2/\beta}+1}.
 \end{align*}

To bound $R_2(T)$, we have
\begin{equation}
\label{eqn:R2Bound1}
    \begin{array}{lcl}
    R^E_2(T)
    &\leq & \Delta_{max}\sum^{l_0}_{l=l^\prime}e^l \delta(l)\\
    &= & \Delta_{max}\sum^{l_0}_{l=l^\prime}e^lPr\{ |\hat{M}(l)-M|>\epsilon(l)/2N  \}\\
    & \leq & \Delta_{max}3NK\sum^{l_0}_{l=l^\prime}e^{l}e^{-\epsilon(l)^2\sum_{i}^l T_s(i)/8N^2},
\end{array}
\end{equation}
where the last inequality follows by applying Hoeffding's inequality on each component of the matrix difference and noting  each is estimate calculated with $\sum_{i}^l T_s(i)$ samples, and taking union bound over all components.
We lower bound the sum $\sum_{i=1}^{l}T_s(i)$ as follows 
\begin{equation*}
    \begin{array}{lcl}
    \sum_{i=1}^{l}T_s(i) 
    &=& 16N^2\sum_{i=1}^{l}i^{(\beta)} \geq 16N^2\int_{x=1}^{l+1} (x-1)^{\beta} \\
    &\geq& 16N^2 (0.5l^{(1+\beta)}).
   \end{array}
 \end{equation*}
Using this in  the last step of (\ref{eqn:R2Bound1}), we get    
\begin{equation*}
    \begin{array}{clc}
  R^E_2(T)
    &\leq &
    3NK\Delta_{max}\sum^{l_0}_{l=l^\prime}e^{l}e^{-\epsilon(l)^2T_s(l)/8N^2}\\
    & \leq & 3NK\Delta_{max}\sum^{l_0}_{l^\prime}e^le^{-\epsilon(l)^2 l^{1+ \beta}}\\
    & = & 3NK\Delta_{max}\sum^{l_0}_{l^\prime}e^le^{-l^{-\beta} l^{1+ \beta}}\\
  & = & 3NK\Delta_{max}\sum^{l_0}_{l^\prime+1}e^le^{-l}\\
  & = & 3NK\Delta_{max}l_0 \leq 3NK\Delta_{max}\log T.
\end{array}
\end{equation*}
The claimed bound follows by adding the bounds on $R^O(T), R^S(T)$ and $R^E(T)$.
\hfill \IEEEQED

\subsection{{\bf Proof of Theorem \ref{thm:mESER}}}
As in the previous proof we bound regret in the exploration, signaling and exploitation phases denoted $R^{O}(T),R^{S}(T)$, and $R^{E}(T)$, respectively. 

\noindent
The values sequences $\{\epsilon(l)\}, \{T_s(l)\}, \{T_b(l)\}$ are set as $\epsilon(l)=l^{-\beta/2}, T_s(l)=16N^2/\epsilon(l)^2$ and $T_b(l)=\lceil \log_2 4N/\epsilon(l) \rceil$ for all $l$. 
 For a given $T$, let $l_0$ denote the phase in which the algorithm terminates. We have $T\geq \sum_{l=1}^{l_0}(T_s(l) + T_b(l) + e^{l})\geq e^{l_0}$ and $\log(T) \geq l_0$.
 
 \noindent
 When the condition in line ($11$) is met, $\epsilon(l)$ is set to a constant in the subsequent epochs. Let $l^\prime$ be the smallest integer such that the condition $\hat{\Delta}_{\min} (l)> 2 \epsilon(l)$ holds. Since $\Delta_{\min} \geq \hat{\Delta}_{\min} (l)- \epsilon(l)$ from Lemma \ref{lma:secondAlloc} (provided $|M-\hat{M}|\leq \epsilon(l)/2N$), we have $\Delta_{
\min}> \epsilon(l^\prime)$, and $l^\prime $ is given by $l^\prime=\lceil \Delta_{\min}^{-2/\beta}\rceil$. For all $l > l^\prime$, $\epsilon(l)=\epsilon(l^\prime), T_s(l)=T_s(l^\prime)$ and $T_s(l)=T_s(l^\prime)$.

\noindent
Regret from the exploration phases is bounded as follows:
\begin{equation*}
\begin{array}{lcl}
R^O(T) 
&\leq &\sum_{l=1}^{l=l^\prime-1} \Delta_{max} T_s(l)K + \sum_{l=l^\prime }^{l=l_0} \Delta_{max} T_s(l)K\\
& \leq & K\Delta_{max} (l^\prime-1) T_s(l^\prime) +K\Delta_{max} T_s(l^\prime) l_0 \\
& \leq & 16N^2K\Delta_{max} (l^\prime)^\beta \left \{(l^\prime-1) + \log T\right\} \\
& \leq & 16N^2K\Delta_{max}(\Delta_{\min}^{-2/\beta}+1)^\beta\left\{\Delta_{\min}^{-2/\beta} + \log T\right \}. 
\end{array}
\end{equation*}
Regret from the signalling phases is bounded as follows: 
\begin{equation*}
\begin{array}{lcl}
R^{S}(T) 
& \leq &  \sum_{l=1}^{l=l_0}T_b(l) KN^2  \\
& = & N^2K \sum_{l=1}^{l=l^\prime-1}T_b(l) + N^2K \sum_{l=l^\prime}^{l=l_0}T_b(l) \\
& \leq & N^2KT_b(l^\prime)(l^\prime-1)+ N^2KT_b(l^\prime)l_0 \\
& \leq & N^2K \left(\log_2 8N + \frac{\beta}{2}\log_2(\Delta_{\min}^{-2/\beta}+1)\right)  \left \{ \Delta_{\min}^{-2/\beta} + \log T \right\}. 
\end{array}
\end{equation*}
We now look at the regret in the exploitation phase. In the $l^{th}$ epoch, the expected regret is bounded as $R^E(l) \leq e^{l}(\epsilon(l)(1-\delta(l)) + \Delta_{max}\delta(l))$. Summing over all regret the epochs we have
\begin{equation*}
    \begin{array}{lcl}
    R^{E}(T) 
    & \leq & \sum^{l^\prime-1}_{l=1}e^{l}(\epsilon(l)+\delta(l)(\Delta_{max}-\epsilon(l))) \\
    && \hspace{50pt}+\Delta_{max}\sum^{l_0}_{l=l^\prime}e^{l}\delta(l)\\
    & = & R^{E}_1(T) + R^{E}_2(T)
    \end{array}
\end{equation*}
where
\begin{equation*}
\begin{array}{lcl}
    R^{E}_1(T) 
    & = & \sum^{l^\prime-1}_{l=1}e^{l}(\epsilon(l)
    +\delta(l)(\Delta_{max}-\epsilon(l)))\\
    & \leq & e^{l^\prime}(\Delta_{max}) = \Delta_{max}e^{1/\Delta_{\min}^{2/\beta}+1},
\end{array}
\end{equation*}    
\begin{equation*}
\begin{array}{lcl}
    R^E_2(T)
    &= & \Delta_{max}\sum^{l_0}_{l=l^\prime}e^lPr\{ |\hat{M}(l)-M|>\epsilon(l^\prime)/2N  \}\\
    & \leq & \Delta_{max}\sum^{l_0}_{l=l^\prime}e^{l}3NKe^{-\epsilon(l^\prime)^2\sum_{j=1}^lT_s(j)/8N^2}\\
          & \leq & 3NK\Delta_{max}\sum^{l_0}_{l=l^\prime}e^le^{-\epsilon(l^\prime)^2 l^{\beta+1}}\\
          & =& 3NK\Delta_{max}\sum^{l_0}_{l=l^\prime}e^le^{-(l^\prime)^{-\beta} (l)^{\beta+1}}\\
    & = & 3NK\Delta_{max}\sum^{l_0}_{l=l^\prime}e^{(l/l^\prime)^{-\beta}} \leq 3NK\Delta_{max}\log T\\
\end{array}
\end{equation*}
where we used Hoeffding's inequality in the 2nd step and  the inequality $\sum_{j=1}^{l}T_s(j) \geq 16N^2 (0.5 (l)^{1+\beta})$ in the rth step. The final regret bound follows by summing the bounds on $R_0(T), R^S(T),$ and $R^E(T).$ \hfill \IEEEQED
\subsection{{\bf Proof of Theorem \ref{thm:mESER_homo}}}

    
    The values sequences $\{\epsilon(l)\}, \{T_s(l)\}, \{T_b(l)\}$ are set as $\epsilon(l)=l^{-\beta/2}, T_s(l)=4/\epsilon(l)^2$ and $T_b(l)=N\lceil\log_2K\rceil +1 $ for all $l$. 
 For a given $T$, let $l_0$ denote the phase in which the algorithm terminates. As earlier, we have $\log(T) \geq l_0$.
 
Let $l^\prime$ be the smallest integer when the condition $\hat{\Delta}_{\min}(l)$ in line $9$ holds. Since $\Delta_{\min} \geq \hat{\Delta}_{\min}(l)-\epsilon$ holds from Lemma \ref{lma:deltamin_homo} (provided $|\mu- \hat{\mu}|\leq \epsilon(l)/2$ at the Leader), $l^\prime$ is such that $\Delta_{\min}\geq \epsilon(l^\prime)$ and we set $l^\prime=\lceil \Delta^{-2/\beta} \rceil$. For all $l> l^\prime$, we have $\epsilon(l)=\epsilon(l^\prime)=(l^\prime)^{-\beta/2}$ and $T_s(l)=T_s(l^\prime)=4(l^\prime)^\beta$.
 
We first the regret from the exploration phase
    \begin{equation*}
    \begin{array}{lcl}
    R^O(T) 
    &\leq &\sum_{l=1}^{l=l^\prime-1} \Delta_{max} T_s(l)K + \sum_{l=l^\prime}^{l=l_0} \Delta_{max} T_s(l_1)K\\
    & \leq & \Delta_{max}K (l^\prime -1)T_s(l^\prime) +\Delta_{max}K l_0 T_s(l^\prime) \\
    & \leq & 4\Delta_{max}K (\Delta_{\min}^{-2/\beta}+1)^{\beta}\left\{\Delta_{min}^{-2/\beta}+ \log(T)\right\}  \\
    \end{array}
    \end{equation*}
    \\    
    For the packetized signaling, $T_b(l) = N\lceil log_2K \rceil + 1$ for all $l\geq 1$. Thus, regret in the signaling phase is  
    \begin{equation*}
    \begin{array}{lcl}
    R^{S}(T) 
    & \leq &  \sum_{l=1}^{l=l_0}T_b(l)  \\
    & \leq & \sum_{l=1}^{l=l_0}N \log_2 K+N+1 \\
    &  \leq & (N \log_2 K+N+1)\log T
    \end{array}
    \end{equation*}
    \\
    We now look at the regret in the exploitation phase. Over the $l^{th}$ epoch, the expected regret would be $R^E(l) \leq e^{l}( \epsilon(l)(1-\delta(l)) + \Delta_{max}\delta(l))$. Thus the overall regret from the exploitation phase is $R^{E}(T) = \sum^{l_0}_{l=1}R^E(l)$ would be:
    \begin{equation*}
        \begin{array}{lcl}
        R^{E}(T) 
        & \leq & \sum^{l^\prime-1}_{l=1}e^{l}(\epsilon(l)+\delta(l)(\Delta_{max}-\epsilon(l))) \\
        &+& \hspace{.5cm} \Delta_{max}\sum^{l_0}_{l=l^\prime}e^{l}\delta(l)\\
        & = & R^{E}_1(T) + R^{E}_2(T).
 \end{array}
    \end{equation*}
    where
  \begin{equation*}
        \begin{array}{lcl}
        R^{E}_1(T) 
        & = & \sum^{l^\prime-1}_{l=1}e^{l}(\epsilon(l)+\delta(l)(\Delta_{max}-\epsilon(l)))\\
        & \leq & e^{l^\prime+1}\Delta_{max} =
        \Delta_{\max}e^{1/\Delta_{\min}^{2/\beta}+1},
         \end{array}
    \end{equation*}
  \begin{equation*}
        \begin{array}{lcl}
        R^E_2(T)
        &= & \Delta_{max}\sum^{l_0}_{l=l_1}e^lPr\{ |\hat{\mu}(l)-\mu|>\epsilon(l^\prime)/2  \}\\
        &= & \Delta_{max}\sum^{l_0}_{l=l^\prime}e^l\Pr( \cap_{j=1,..K} | \hat{\mu}_{j}(l) - \mu_{j} | > \epsilon(l^\prime)/2 )\\
         & \leq & \Delta_{max}\sum^{l_0}_{l=l^\prime}e^l \sum_{j=1}^{j=K} \Pr(| \hat{\mu}_{j}(l) - \mu_{j} | > \epsilon(l^\prime)/2) \\
        &\leq &\Delta_{max}\sum^{l_0}_{l=l^\prime}e^l \sum_{j=1}^{j=K} 2e^{-2\epsilon(l^\prime)^2 \sum_{i=1}^l T_s(i)/4} \\
         &\leq &\Delta_{max}\sum^{l_0}_{l=l^\prime}e^l \sum_{j=1}^{j=K} 2e^{-(l^\prime)^{-\beta} (l)^{\beta+1}}\\
          &\leq &\Delta_{max}\sum^{l_0}_{l=l^\prime}2Ke^{(l/l^\prime)^{-\beta}} \leq 2K \Delta_{\max}\log T\\
       
    \end{array}
    \end{equation*}
where we applied uninon bound in the first inequlity, Hoeffding's bound in the 2nd inequality and the relation $\sum_{i=1}^lT
_s(i)\geq 4 (0.5 l^{\beta+1})$ in the 3rd inequality. The final bound follows by summing the bound on $R^0(T), R^S(T),$ and $R^E(T)$.
     \hfill \IEEEQED

\section*{Wideband sensing}
\label{sec:Wide-band Sensing}
In this subsection, we develop an algorithm named  Distributed Optimal Assignment with Wideband Sensing (DOA-WS). The inputs to the algorithm are $K,T_s,T_r$ and $T_b$ and the algorithm guarantees an $(\epsilon, \delta)$-optimal assignment within a finite number of rounds, where $\epsilon$ and  $\delta$ are dependent on $K,T_s,T_r$ and $T_b$. The algorithm consists of three phases namely 1) Random Hopping (RH) 2) Sequential Hopping (SH) and 3) Bernoulli Signaling (BS). These phases run sequentially to estimate the matrix $M$. In the end, all players apply the Hungarian method\footnote{The Hungarian algorithm gives an assignment that minimizes the sum reward. Since we want the sum reward to be maximized, the input given to this algorithm is the negative of the rate matrix, thus ensuring the sum reward is maximized. Henceforth, it is assumed implicitly that the negative of the rate matrix is provided to the algorithm.} and play the arm given by the optimal assignment thereafter. All subroutines are written in a distributed fashion. The pseudo-code of DOA-WS is given in Algorithm (\ref{alg:Algo1}). The algorithm is run by each player separately. We suppress index $n$ to avoid cluttering in the presentation.


\begin{algorithm}[!h]
	\caption{DOA-WS} \label{alg:Algo1}
	\begin{algorithmic}[1]
		\State	Input: {$K, T_s, T_r, T_b$}
		\State	Initial Orthogonal Assignment: $(N, k,n)= RH(K, T_r)$ 
		\State	 \textbf{Exploration}: $\hat{\mu}_n=SH(K, k, T_s )$ 
		\State  \textbf{Signaling}: $\hat{M}=BS(N,\hat{\mu}_n, T_b)$
		\State Find  optimal assignment $\hat{\pi}^*$ on $\hat{M}$ 
		\State \textbf{Exploitation}: Play $\hat{\pi}^*(k)$ till end
	\end{algorithmic}
\end{algorithm}
The RH phase ensures that all players are on different arms (orthogonalized). During each round of this phase, each player selects an arm uniformly at random until they experience a collision-free round. Once it happens, they continue playing that arm till the end of the phase. The length of the phase is set such that all players are orthogonalized with high probability.
\begin{algorithm}[!h]
	\caption*{\textbf{Phase 1:} Random Hopping (RH)}
	\begin{algorithmic}[1]
		\State Input: $K,T_r$  
		\State Initialize: Set $Lock=0$
		\For{$t=1 \dots T_{r}$}
		\If{($Lock == 1$)}
		\State Select the same arm, $\alpha_t = \alpha_{t-1}$
		\Else
		\State Randomly select a channel, $\alpha_t \sim U([K])$
		\State Set $Lock=1$ if no collision is observed
		\EndIf
		\EndFor	
		\State Set $k=\alpha_t$. Transmit on $k$ and sense all other arms
		\State $A=\left \{i\in [K]: \mbox{transmission is sensed on arm } i \neq k \right \}$
		\State Set $N = |A|+1$ and $n = |\{j \in A : j < k\}| + 1$
		\State Return $N,n$ and index of current arm $k$.
	\end{algorithmic}
\end{algorithm}

The SH phase ensures that all the players learn mean rewards they see on the arms. In this phase, all the players select the arms sequentially i.e. arm $(k+1)\mod K$ is played in the next round after playing arm $k$ in the current round. Since an orthogonal allocation is maintained in each round, no collisions occur in this phase. The length of the phase is set such that the mean rewards are estimated with high accuracy. 

\begin{algorithm}[!h]
	\caption*{\textbf{Phase 2:} Sequential Hopping (SH)}
	\begin{algorithmic}[1]
		\State Input: $K, T_s, k,n$  
		\State Set $\alpha_{T_{r}+K}=k$, $r_i=0 \;\; \forall i\in [K]$
		\For{$t=T_{r}+K+1 \dots KT_{s}+K$}
		\State Play arm $\alpha_t=(\alpha_{t-1} +1 )\mod K$
		\State Observe reward $X_{\alpha_t}$ and update $r_{\alpha_t}=r_{\alpha_t} + X_{\alpha_t}$
		\EndFor	
		\State Estimate $\hat{\mu}_{n,k}=r_{k}/T_{s} \; \forall k \in [K]$
		\State Return $\hat{\mu}_n= \{\hat{\mu}_{n,k}\}$.
	\end{algorithmic}
\end{algorithm}

In the BS phase, each player signals her observation of mean rewards to others and also learns reward seen by others, through their signals. This is achieved as follows: The BS phase consists of $K$ frames each of $T_b$ rounds in which each player plays the same arm for the entire phase. In the $i^{th}$ frame, each player transmits according to a Bernoulli distribution with the Bernoulli parameter set to the value of their estimated mean on arm $i$. They also detect whether transmissions have occurred on other arms and update the number of transmissions they observe on each arm. At the end of $K$-{th} frame, each player computes the mean number of transmissions observed on each arm, including the arm on which she transmitted. The method of update is such that $\hat{M}$ formed is a $K \times K$ matrix. Since there are $N$ players, in every column there will be only $N$ non-zero entries. We remove those $K-N$ zero rows and we are left with the $N \times K$ estimated mean reward matrix.

\begin{algorithm}[!h]
	\caption*{\textbf{Phase 3:} Bernoulli Signaling (BS)}
	\begin{algorithmic}[1]
		\State Input: $K, T_b, \hat{\mu}_n$  
		\State Set $o_{i,j}=0 \;\forall\; i, j\in [K]$  
		\For{$j=1,2\dots, K$}
		\For {$T_b$ number of rounds}
		\State Transmit with probability $\hat{\mu}_{n,j}$ on current arm
		\State $A=\left \{k\in [K]: \mbox{transmission is sensed on arm } k\right \}$
		\State  Update $o_{i,j}= o_{i,j}+ 1$ for all $i \in A$
		\EndFor
		\EndFor	
		\State Estimate $\widehat{\mu}_{i,j}=o_{i,j}/T_b \; \forall \; i,j \in [K]$
		\State Return $\hat{M}= \left \{\widehat{\mu}_{i,j}\right \}$.
	\end{algorithmic}
\end{algorithm}

\noindent
At the end of the BS phase, each player will have the same estimate rate matrix $\hat{M}$. Each player then finds the optimal assignment on $\hat{M}$ and plays it thereafter. We state the following lemmas before providing the guarantees of DOA-WS:
\begin{lemma}
	\label{lma:RHPhase}
	Let $\delta \in (0,1)$. If RH phase is run for  $ T_{r}=\left \lceil \frac{\log(\delta/3K)}{ \log\left(1-1/4K\right)} \right \rceil $ number of rounds then all the players will be orthogonalized with probability at least $1-\delta/3$.  
\end{lemma}  

\begin{lemma}
		\label{lma:SHPhase}
For any given $\delta \in (0,1)$ and $\epsilon>0$ set $T_s = \frac{8N^2}{\epsilon^2}\log\left(\frac{6NK}{\delta}\right)$. Then estimated  mean rewards at the end of the SH phase  are such that $|\mu_{n,k}-\hat{\mu}_{n,k}|\leq \epsilon/4N$ for all $n\in [N], k\in [K]$ with probability at least $1 - \delta/3$.
\end{lemma}

\begin{lemma}
		\label{lma:BSPhase}
	For any given $\delta \in (0,1)$ and $\epsilon>0$ set  $T_b = \frac{8N^2}{\epsilon^2}\log\left(\frac{6K^2}{\delta}\right)$. Let $\hat{\mu}_n$ denotes the mean reward vector that a player $n\in [N]$ signals. Then estimated  mean rewards matrix $\hat{M}=\left \{\widehat{\mu}_{i,j}\right \}$ at the end of the BS phase will be  such that $|\hat{\mu}_{n,k}-\widehat{\mu}_{n,k}|\leq \epsilon/4N$ for all $n\in [N], k\in [K]$ with probability at least $1 - \delta/3$.
\end{lemma}
Proof for Lemma \ref{lma:RHPhase} is similar to \cite{WIOPT2013_TrekkingBased_KumarYadavDarak}[Lemma 1]. The proofs for Lemmas \ref{lma:SHPhase} and \ref{lma:BSPhase} are straightforward applications of the Hoeffding's concentration inequalities \cite{hoeffding1963probability} and have been omitted due to space constraints.

\begin{thm}
	\label{thm:DOA-WS}
	For a given $\epsilon>0, \delta \in (0,1)$ let $T_r, T_s$ and $T_b$ be set as in the Lemmas \ref{lma:RHPhase}, \ref{lma:SHPhase} and \ref{lma:BSPhase} respectively. Then the DOA-WS policy is $(\epsilon, \delta)$-optimal after $T_r+KT_s + KT_b$ number of rounds for any mean reward matrix $M$. 
\end{thm}
\noindent
{\textit{Proof of Theorem (\ref{thm:DOA-WS}):}} If (A) denotes all players orthogonalized in RH phase, (B) denotes all players estimate their mean reward within $\epsilon/4N$ accuracy and (C) denotes all players signal their estimates to others within $\epsilon/4N$ accuracy,  the estimated matrix $\hat{M}$ at each player will be such that $|\hat{M}- M|\leq \epsilon I/2N$ after $T:=T_r + K(T_s+ T_b)$ rounds, where $I$ is an all-ones $N\times K$ matrix. Theorem \ref{thm:ApproxHun} then implies that if all players play the optimal assignment obtained from $\hat{M}$, DOA-WS policy will be $\epsilon$-optimal for all $t > T$. Hence we have
\begin{eqnarray*}
	\lefteqn{\Pr\left \{\mbox{DOA-FS is  $\epsilon$-optimal for}\;\; t > T\right \} \geq \Pr\{\text{A,B,C holds}\} }\\
	&=& 1-\Pr\{\text{any of A,B,C doesn't hold}\}> 1-3\delta/3= 1-\delta,
\end{eqnarray*}
where the inequality follows from Lemmas (\ref{lma:RHPhase})-(\ref{lma:BSPhase}). \hfill\IEEEQED
\begin{remark}
Note that mean reward corresponding to player $n$ in $\hat{M}$ is $\widehat{\mu}_n$ and not $\hat{\mu}_n$, i.e., the values that players signaled to other players in the BS phase and not what she estimated in the SH phase. This is done to ensure that all players have the same $\hat{M}$ so that they find the same optimal assignment.
\end{remark}

\begin{remark}
The performance in the BS phase can be improved by listening only on the arms on which players are transmitting. In this case, we can set 
$T_b = \frac{8N^2K}{\epsilon^2}\log\left(\frac{6NK}{\delta}\right)$ and still achieve the same result. We skip the details.
\end{remark}

\section*{Packetized vs Bernoulli Signaling }
In the BS phase of DOA-WS, players exchange information with high accuracy using Bernoulli distributions. This can be achieved in multiple other ways. One possibility is that each player encodes her estimates and signals this
using a certain number of bits. For example, each player can encode her estimates within $\epsilon/4N$ accuracy using $\lceil \log_2 (4N/\epsilon)\rceil$ bits. Then, each player needs $K\lceil \log_2 (4N/\epsilon)\rceil$ bits to encode her estimates of all arms. The players can then transmit on an arm as per the bit sequence of their code -- a transmission corresponds to bit '$1$' and a no transmission corresponds to bit  '$0$'. Since players can sense all the arms, within $K\lceil \log_2 (4N/\epsilon )\rceil$ time slots, all players will receive the estimates of other players within $\epsilon/4N$ accuracy. Theorem \ref{thm:DOA-WS} will hold for packetized signaling as well, by setting $T_b=K\lceil \log_2 (4N/\epsilon )\rceil$.

Packetized signaling requires less number of time slots compared to Bernoulli signaling for information exchange. However, it is more prone to errors. An error in a single bit, especially if it is a leading bit, can cause significant distortion in the information exchanged by packetized signaling,  whereas Bernoulli signaling is more robust to such distortions. The choice of a signaling scheme depends on sensing accuracy of players and how robustly signaling can be done.

\section*{Acknowledgments}
 Manjesh K. Hanawal would like to thank support from INSPIRE
 faculty fellowships from  DST, Government of India, SEED grant (16IRCCSG010) from
 IIT Bombay, and Early Careers Research Award (ECRA) from SER    B. Sumit J Darak would like to thank support from INSPIRE faculty fellowship from DST, Government of India. 

We thank Abbas Mehrabian and Emilie Kaufmann for pointing to inaccuracies in the conference version of this draft.

\bibliographystyle{IEEEtran}
\bibliography{biblio}

\begin{thebibliography}{10}
\providecommand{\url}[1]{#1}
\csname url@samestyle\endcsname
\providecommand{\newblock}{\relax}
\providecommand{\bibinfo}[2]{#2}
\providecommand{\BIBentrySTDinterwordspacing}{\spaceskip=0pt\relax}
\providecommand{\BIBentryALTinterwordstretchfactor}{4}
\providecommand{\BIBentryALTinterwordspacing}{\spaceskip=\fontdimen2\font plus
\BIBentryALTinterwordstretchfactor\fontdimen3\font minus
  \fontdimen4\font\relax}
\providecommand{\BIBforeignlanguage}[2]{{%
\expandafter\ifx\csname l@#1\endcsname\relax
\typeout{** WARNING: IEEEtran.bst: No hyphenation pattern has been}%
\typeout{** loaded for the language `#1'. Using the pattern for}%
\typeout{** the default language instead.}%
\else
\language=\csname l@#1\endcsname
\fi
#2}}
\providecommand{\BIBdecl}{\relax}
\BIBdecl

\bibitem{JSAC11_DistributedLearning_Anadakumar}
A.~Anandkumar, N.~Michael, A.~K. Tang, and A.~Swami, ``Distributed algorithms
  for learning and cognitive medium access with logarithmic regret,''
  \emph{IEEE Journal on Selected Areas in Communications}, vol.~29, no.~4, pp.
  731--745, 2011.

\bibitem{TMC11_CongnitiveMediumAccess_LaiGamalJiangPoor}
L.~Lai, H.~E. Gamal, H.~Jiang, and H.~V. Poor, ``Cognitive medium access:
  Exploration, exploitation, and competition,'' \emph{IEEE Transaction on
  Mobile Computing}, vol.~10, no.~2, 2011.

\bibitem{Book12_RegretAnalysis_Bubeck}
S.~Bubeck and N.~Cesa-Bianchi, \emph{Regret analysis of stochastic and
  non-stochastic multi-armed bandit problems}.\hskip 1em plus 0.5em minus
  0.4em\relax {NOW} publisher, Foundations and Trends in Machine Learning,
  2012.

\bibitem{NRL1955_TheHingarianMethod_Kuhn}
H.~W. Kuhn, ``The hungarian method for the assignment problem,'' \emph{Naval
  Research Logistics (NRL)}, vol.~2, no. 1-2, pp. 83--87, 1955.

\bibitem{IAT1957_AlgorithmForTheAssignment_Munkres}
J.~Munkres, ``Algorithms for the assignment and transportation problems,''
  \emph{Journal of the Society for Industrial and Applied Mathematics}, vol.~5,
  no.~1, pp. 32--38, 1957.

\bibitem{JMLR2006_ActionEliminationAndStoppingConditions_DarMannorMansour}
E.~Even-Dar, S.~Mannor, and Y.~Mansour, ``Action elimination and stopping
  conditions for the multi-armed bandit and reinforcement learning problems,''
  \emph{Journal of Machine Learning and Research}, vol.~7, pp. 1079--1105,
  2006.

\bibitem{JMLR2004_TheSampleComplexityOfExploration_MannorTsitsiklis}
S.~Mannor and J.~N. Tsitsiklis, ``The sample complex- ity of exploration in the
  multi-armed bandit problem,'' \emph{Journal of Machine Learning and
  Research}, vol.~6, pp. 623 -- 648, 2004.

\bibitem{TSP10_DistributedLearning_LiuZhao}
K.~Liu and Q.~Zhao, ``Distributed learning in multi-armed bandit with multiple
  players,'' \emph{IEEE Transactions on Signal Processing}, vol.~58, no.~11,
  2010.

\bibitem{WC2016_DistributedStochchasticLearning_ZhadiDongGrami}
M.~Zandi, M.~Dong, and A.~Grami, ``Distributed stochastic learning and
  adaptation to primary traffic for dynamic spectrum access,'' \emph{IEEE
  Transaction on Wireless Comms}, vol.~15, no.~3, pp. 1675--1688, 2016.

\bibitem{TSP14_DistributedStochastic_GaiKrishnamachari}
Y.~Gai and B.~Krishnamachari, ``Distributed stochastic online learning policies
  for opportunistic spectrum access,'' \emph{IEEE Transactions on Signal
  Processing}, vol.~62, no.~23, 2014.

\bibitem{ICML16_MultiplayerBandits_RosenkiShamir}
J.~Rosenski, O.~Shami, and L.~Szlak, ``Multi-player bandits -- a musical chairs
  approach,'' in \emph{Proceedings of International Conference on Machine
  Learning (ICML)}, New York, USA, 2016.

\bibitem{KDD14_ConcurrentBandits_AvnerMannor}
O.~Avner and S.~Mannor, ``Concurrent bandits and cognitive radio networks,'' in
  \emph{Proceedings of the Machine Learning and Knowledge Discovery in
  Databases}.\hskip 1em plus 0.5em minus 0.4em\relax Stanford, CA: Springer,
  2014.

\bibitem{WIOPT2013_TrekkingBased_KumarYadavDarak}
R.~Kumar, A.~Yadav, S.~J. Darak, and M.~K. Hanawal, ``Trekking based
  distributed algorithm for opportunistic spectrum access in infrastructureless
  network,'' in \emph{16th International Symposium on Modeling and Optimization
  in Mobile, Ad Hoc, and Wireless Networks (WiOpt)}, 2018.

\bibitem{TIT14_DecentralizedLearning_KalthilNayyarJain}
D.~Kalathil, N.~Nayyar, and R.~Jain, ``Decentralized learning for multiplayer
  multiarmed bandits,'' \emph{IEEE Transactions on Information Theory},
  vol.~60, no.~4, pp. 2331--2345, 2014.

\bibitem{INFOCOM16_MultiUserLax_AvnerMannor}
O.~Avner and S.~Mannor, ``Multi-user lax communications: A multi-armed bandit
  approach,'' in \emph{IEEE International Conference on Computer Communications
  (INFOCOM)}, San Francisco, CA, USA, 2016.

\bibitem{TCNS2018_DecentralizedLearning_KalthilNayyarJain}
N.~Nayyar, D.~Kalathil, and R.~Jain, ``On regret-optimal learning in
  decentralized multiplayer multiarmed bandits,'' \emph{IEEE Transactions on
  Control of Network Systems}, vol.~5, no.~1, pp. 597--606, 2018.

\bibitem{INFOCOM2018_Low-ComplexityLearning_KangJoo}
S.~Kang and C.~Joo, ``Low-complexity learning for dynamic spectrum access in
  multi-user multi-channel networks,'' in \emph{IEEE International Conference
  on Computer Communications (INFOCOM)}, 2018.

\bibitem{GoT}
I.~Bistritz and A.~Leshem, ``Distributed multi-player bandits-a game of thrones
  approach,'' in \emph{Advances in Neural Information Processing Systems},
  2018, pp. 7221--7231.

\bibitem{kaufmann2019new}
E.~Kaufmann and A.~Mehrabian, ``New algorithms for multiplayer bandits when arm
  means vary among players,'' \emph{arXiv preprint arXiv:1902.01239}, 2019.

\bibitem{Infocom2019_DistributedLearning_TibrewalPatchalaHanawal}
M.~K.~H. Harshvardhan~Tibrewal, Sravan~Patchala and S.~J. Darak, ``1distributed
  learning and optimal assignment inmultiplayer heterogeneous networks,'' in
  \emph{to appear in IEEE International Conference on Computer Communications
  (INFOCOM)}, 2019.

\bibitem{hoeffding1963probability}
W.~Hoeffding, ``Probability inequalities for sums of bounded random
  variables,'' \emph{Journal of the American statistical association}, vol.~58,
  no. 301, pp. 13--30, 1963.

\bibitem{Scipy_HungarianAlgorithm}
B.~M. Clapper and G.~Varoquaux, ``{Hungarian algorithm (Kuhn-Munkres) for
  solving the linear sum assignment},''
  \url{github.com/scipy/scipy/blob/v0.18.1/scipy/optimize/$\_$hungarian.py/$\#$L13-L107},
  2016, [Online; accessed 30-July-2018].

\bibitem{ALT2018_MultiplayerBandits_BessonKaufma}
L.~Besson and E.~Kaufmann, ``Multi-player bandits models revisited,'' in
  \emph{To appear in Algorithmic Learning Theory (ALT)}, 2018.

\bibitem{matsui1994algorithms}
T.~Matsui, A.~Tamura, and Y.~Ikebe, ``Algorithms for finding a kth best valued
  assignment,'' \emph{Discrete Applied Mathematics}, vol.~50, no.~3, pp.
  283--296, 1994.

\bibitem{avner2018multi}
O.~Avner and S.~Mannor, ``Multi-user communication networks: A coordinated
  multi-armed bandit approach,'' \emph{arXiv preprint arXiv:1808.04875}, 2018.

\end{thebibliography}


\end{document}